%% file: iclr2027_conference.tex
\documentclass{article}
\PassOptionsToPackage{table}{xcolor}
\usepackage{iclr2027_conference}
\usepackage{times}
\usepackage[T1]{fontenc}

\usepackage{amsmath,amssymb}
\usepackage{graphicx}
\usepackage{booktabs}
\usepackage{multirow}
\usepackage{wrapfig}
\usepackage{xcolor}
\usepackage{hyperref}
\usepackage{url}
\definecolor{paperblue}{HTML}{356A93}
\definecolor{paperteal}{HTML}{258779}
\definecolor{papertint}{HTML}{EDF5F3}
\hypersetup{
    colorlinks=false,
    pdfborder={0 0 1},
    citebordercolor={0 1 0},
    pdftitle={VaME: Exploring Variational Latent Reasoning for Multimodal Embeddings},
    pdfauthor={Peixi Wu; Mingzhou Jiang; Feipeng Ma; Biao Yang; Yunhao Zhou;
        Wei Yuan; Bosong Chai; Huizu Lin; Jie Chen; Zhangchi Hu; Fan Yang;
        Wenwu Ou; Hebei Li; Xiaoyan Sun}}
\newcommand{\method}{VaME}
\newcommand{\norm}[1]{\operatorname{Norm}\!\left(#1\right)}

\newcommand{\KL}{D_{\mathrm{KL}}}

\title{VaME: Exploring Variational Latent\\Reasoning for Multimodal Embeddings}
\author{
Peixi Wu$^1$\thanks{\fontsize{8}{10}\selectfont
This work was done during an internship at Kuaishou.\enspace
\textdagger\,Project leader.\enspace\textdaggerdbl\,Corresponding author.},
Mingzhou Jiang$^2$, Feipeng Ma$^1$, Biao Yang$^{3,\dagger}$,
Yunhao Zhou$^3$,\\
\textbf{Wei Yuan$^3$, Bosong Chai$^4$, Huizu Lin$^1$, Jie Chen$^1$, Zhangchi Hu$^1$,}\\
\textbf{Fan Yang$^3$, Wenwu Ou$^3$, Hebei Li$^{1,\ddagger}$, Xiaoyan Sun$^1$}\\
$^1$University of Science and Technology of China\\
$^2$Tsinghua University, $^3$Kuaishou Technology, $^4$Zhejiang University\\
\texttt{\{wupeixi,lihebei\}@mail.ustc.edu.cn, sunxiaoyan@ustc.edu.cn}
}

\iclrfinalcopy 

\begin{document}

\maketitle
\pagestyle{plain} 

\begin{abstract}
\input{sections/abstract}
\end{abstract}

\input{sections/introduction}
\input{sections/related_work}
\input{sections/preliminaries}
\input{sections/method}
\input{sections/experiments}
\input{sections/deep_analysis}
\input{sections/conclusion}

\input{sections/statements}
\bibliography{references}
\bibliographystyle{iclr2027_conference}
\clearpage
\appendix
\input{sections/appendix}
\end{document}

%% file: sections/abstract.tex
Universal multimodal retrieval requires compact embeddings that preserve
task-relevant semantic information across diverse modalities.
Prior works have incorporated latent reasoning into multimodal embedding
learning to refine this information before embedding extraction.
However, most existing approaches remain confined to deterministic latent
paths, without exploring alternative trajectories to discover better
embeddings. Thus, we propose \textbf{\method{}}
(\emph{\textbf{Va}riational \textbf{M}ultimodal \textbf{E}mbeddings}), a framework that models latent
reasoning as a learnable distribution over trajectories.
Specifically, we first introduce Variational Latent Reasoning (VLR) to enable
autoregressive exploration in latent space, guided by answer reconstruction
through a lightweight decoder.
Meanwhile, we augment the original embedding-token readout with a latent-fused
embedding to facilitate exploration during subsequent reinforcement learning.
Finally, we optimize latent reasoning over stochastic variational trajectories
through reinforcement learning, using Semantic Decoding Reward (SDR) to favor
semantically meaningful trajectories with interpretable decoded outcomes.
On the 78-task MMEB-V2 benchmark, spanning image, video, and visual-document
retrieval, VaME outperforms most explicit CoT-based models and
all latent-reasoning baselines.
VaME also demonstrates robust performance on reasoning-intensive benchmarks
such as MRMR, with substantial gains after reinforcement learning.
Importantly, VaME achieves these gains with at least a $\boldsymbol{4.25\times}$ inference
speedup over the deterministic latent autoregressive baselines.
\textit{The code will be made publicly available.}
%

%% file: sections/introduction.tex
\section{Introduction}
\label{sec:intro}

Universal Multimodal Embedding (UME) aims to map text, images, videos, visual
documents, and their mixtures into a shared embedding space, allowing a single
model to serve diverse retrieval tasks~\citep{clip,uniir}. Recent vision-language embedding
models~\citep{lamra,mme5} have made substantial progress~\citep{vlm2vec,gme} by learning instruction-aware
representations~\citep{mmembed,e5v} across these modalities~\citep{vlm2vecv2,llave}.
Yet universal retrieval is not simply a matter of compressing more modalities
into one vector~\citep{metaembed}. Depending on the instruction, the model must identify and compose task-relevant
evidence into a compact representation.

Accordingly, a growing line of work augments embedding formation with
intermediate reasoning. Explicit approaches make reasoning human-readable
through retrieval-oriented rewrites~\citep{rime,mmembr1} or textual reasoning
traces~\citep{umer1,embedrl,tte}, but autoregressive generation~\citep{tte,tteflash}
makes embedding extraction far slower than direct extraction, limiting
large-scale, real-time use. Latent reasoning offers a compact
alternative~\citep{codi,plume}: LaME~\citep{lame} and TTE-Flash~\citep{tteflash}
reason in hidden space while retaining efficient retrieval.
\mbox{Most latent approaches}~\citep{coconut,codi,plume} remain deterministic.
\mbox{Each input follows} one path that can refine its representation but cannot
explore alternative trajectories~\citep{softcotpp}. Training therefore cannot
compare different latent computations for the same input, limiting
exploration-based learning. This raises our central question: \emph{Can an
embedding model explore diverse latent reasoning paths while retaining
efficient inference?}

\begin{figure}[t]
    \centering
    \includegraphics[width=\linewidth]{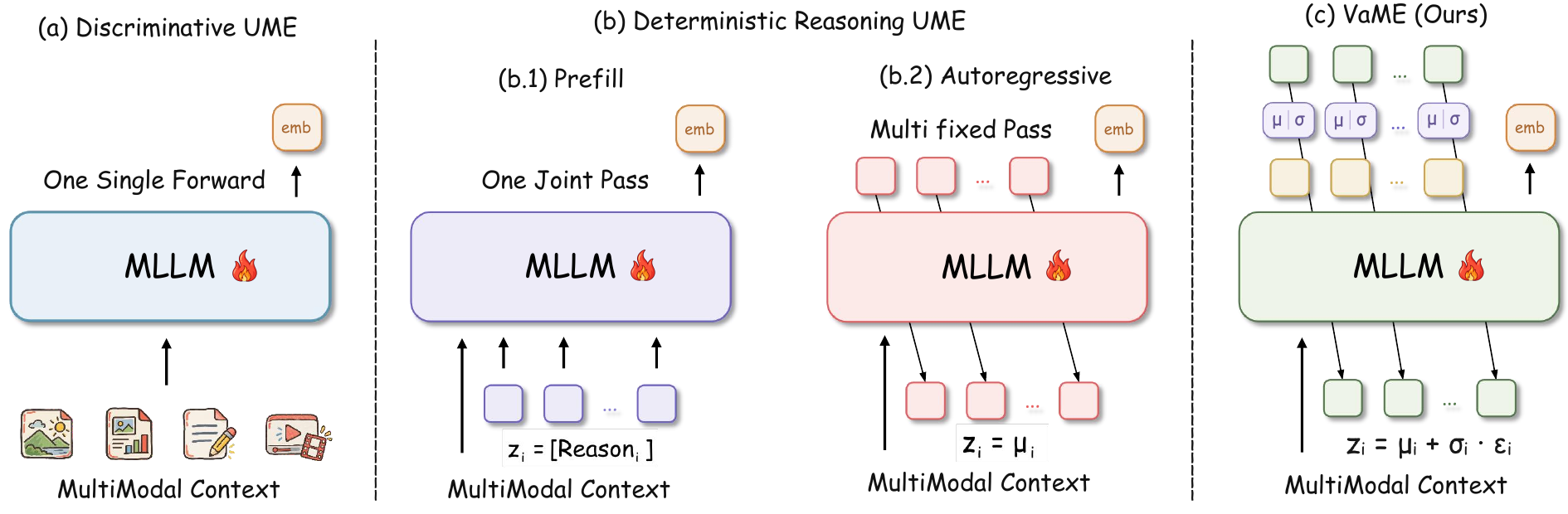}
    \caption{\textbf{From deterministic embedding to variational latent reasoning.}
    (a) Discriminative embedders map each input directly to a single embedding.
    (b) Deterministic reasoning adds latent computation but follows one fixed path.
    (c) \method{} samples and compares alternative latent trajectories during
    training and follows the conditional means for deterministic inference.}
    \label{fig:progression}
\end{figure}

To answer this question, we propose \textbf{\method{}}
(\textbf{Va}riational \textbf{M}ultimodal
\textbf{E}mbeddings), which learns a distribution over latent reasoning
trajectories (Figure~\ref{fig:progression}). Variational Latent Reasoning (VLR)
replaces deterministic transitions with conditional Gaussians~\citep{vae,rezende}.
Each autoregressive step feeds a sampled continuous latent action back into
the backbone, changing the context for subsequent actions~\citep{coconut,codi}.
During latent supervised fine-tuning, retrieval objectives align trajectories
with relevance, while a
lightweight decoder reconstructs answer annotations from their hidden
states~\citep{cafe}. Alongside variational regularization~\citep{vib,climb},
this supervision helps preserve retrieval-relevant semantics without intermediate
chain-of-thought supervision.

To balance exploration and retrieval stability, we retain the original
embedding-token readout for stable single-vector retrieval and add a complementary
latent-fused embedding for trajectory-sensitive retrieval supervision and
reinforcement learning. Attention pooling summarizes the latent states before
residual fusion with the embedding-token state. In the second stage, we sample
multiple variational
trajectories~\citep{colar} and refine the continuous latent policy through
group-relative optimization~\citep{grpo}. Retrieval feedback rewards
discriminative trajectories~\citep{rime,embedrl}, while Semantic Decoding Reward
(SDR) evaluates whether the frozen lightweight decoder produces answers
semantically consistent with the reference. SDR discourages semantically empty
exploration and makes latent reasoning outcomes inspectable through decoding.
At inference, VaME follows one mean trajectory and emits one embedding per
item, without text generation or trajectory search.

Across the 78 tasks of MMEB-V2~\citep{vlm2vecv2}, covering image, video, and visual
document retrieval, VaME outperforms most explicit CoT-based models~\citep{umer1,tte,embedrl} and all
comparable prior latent-reasoning baselines~\citep{lame,tteflash,plume}. It further demonstrates robust performance
on reasoning-intensive benchmarks such as MRMR~\citep{mrmr}, with substantial gains after
reinforcement learning. Importantly, VaME achieves these gains with at least a $\boldsymbol{4.25\times}$ inference
speedup over the deterministic latent autoregressive baselines.
Our contributions are as follows:

\begin{itemize}
    \item We introduce Variational Latent Reasoning, which formulates multimodal latent reasoning as an
    autoregressive distribution over continuous trajectories, enabling
    training-time exploration with deterministic single-trajectory inference.
    \item We develop a two-stage framework in which supervised fine-tuning
    aligns variational trajectories with retrieval relevance and answer
    semantics, and reinforcement learning then refines the latent policy using
    retrieval feedback and semantic decoding reward.
    \item Extensive experiments on MMEB-V2 and MRMR demonstrate that VaME
    delivers strong retrieval performance across diverse multimodal retrieval
    tasks while achieving substantially faster inference than deterministic
    latent autoregressive baselines.
\end{itemize}

%% file: sections/related_work.tex
\section{Related Work}
\label{sec:related}

\noindent\textbf{Universal multimodal embeddings.}\ 
Prior works such as VLM2Vec~\citep{vlm2vec} and GME~\citep{gme} train instruction-aware representations~\citep{e5v,mme5,tsembed,unimev2} with vision--language
backbones~\citep{qwen2vl,llave}. MM-Embed~\citep{mmembed} studies universal retrieval~\citep{uniir,unime} across
multimodal inputs~\citep{rzenembed,lamra}, while VLM2Vec-V2~\citep{vlm2vecv2} extends evaluation and
training to video and visual document tasks. MetaEmbed~\citep{metaembed} explores
flexible late interaction~\citep{colbertv2} and the allocation of multiple vectors at test
time. Our focus is complementary: we change the computation
that precedes extraction while retaining a single-vector retrieval interface.


\noindent\textbf{Thinking before embedding.}\ 
A growing line of work introduces an intermediate reasoning stage before deriving the final
retrieval embedding~\citep{codi,implicitcot}. Some methods enrich the context for embedding extraction
through textual CoT~\citep{rime,mmembr1,umer1,embedrl}, but require multi-step autoregressive decoding.
Other methods process learned tokens in a single prefill
pass~\citep{lame,tteflash} or autoregressively feed hidden states
back into the backbone~\citep{plume,tteflash}. CoT generation can further
supervise these latent representations~\citep{tteflash}. These methods rely on
explicit rewrites or deterministic latent computation. We instead use a stochastic policy over
vector transitions, allowing the model to explore different latent reasoning paths.


\noindent\textbf{Continuous reasoning and variational representations.}\ 
Continuous reasoning shifts intermediate computation from text to hidden states.
Coconut~\citep{coconut} feeds hidden states back recurrently, while CoDI~\citep{codi},
CoLaR~\citep{colar}, and SoftCoT~\citep{softcot,softcotpp} learn or compress reasoning in continuous
space. LaSER~\citep{laser} and PLUME~\citep{plume} apply latent reasoning to dense and multimodal
retrieval. In parallel, variational autoencoders model reparameterized latent
variables~\citep{vae,rezende}, while information-bottleneck methods learn stochastic
representations~\citep{ib,vib}, including multimodal and contrastive variants~\citep{omib,climb}.
Our method seeks to introduce exploratory latent reasoning for multimodal retrieval.


%% file: sections/preliminaries.tex
\section{From Prefill to Autoregressive Latent Thinking}
\label{sec:prelim}

\noindent\textbf{Multimodal retrieval formulation.}\ 
Universal multimodal retrieval seeks to identify the most relevant target
from a candidate pool for an instruction-conditioned query. Following standard
practice, we obtain a multimodal embedder by contrastively
fine-tuning a pretrained MLLM. Given a minibatch
$\mathcal B=\{(q_i,t_i)\}_{i=1}^{B}$ containing $B$ matched pairs, let the MLLM
backbone $F_\theta$ encode $q_i$ and $t_i$ into query and target embeddings
$F_\theta(q_i),F_\theta(t_i)\in\mathbb R^d$, respectively, with each given by
the normalized final-layer hidden state of the last token. The corresponding
InfoNCE objective is
\begin{equation}
    \mathcal L_{\mathrm{NCE}}
    =-\frac{1}{2B}\sum_{i=1}^{B}\left[
      \log\frac{e^{F_\theta(q_i)^\top F_\theta(t_i)/\tau}}
      {\sum_j e^{F_\theta(q_i)^\top F_\theta(t_j)/\tau}}
      +\log\frac{e^{F_\theta(t_i)^\top F_\theta(q_i)/\tau}}
      {\sum_j e^{F_\theta(t_i)^\top F_\theta(q_j)/\tau}}
      \right],
    \label{eq:contrastive}
\end{equation}
where $\tau$ denotes the temperature; $\{t_j\}_{j\ne i}$ serve as in-batch
negatives for $q_i$, and vice versa.

\noindent\textbf{From prefill to latent recurrence.}\ 
Reasoning-enhanced embedding models differ in the computation performed before
representation extraction. LaME and register-based TTE-Flash place learnable
latent slots $\{r_k\}_{k=1}^{K}$ in a single prefill pass, preserving
parallelism~\citep{lame,tteflash}. Their hidden states attend to earlier
positions, but each input embedding $r_k$ is shared across examples and does
not depend on the preceding latent state $h_{k-1}$. Recurrence instead takes
$h_{k-1}$ as its step-$k$ input, enabling input-dependent refinement:
\begin{equation}
    (h_0,c_0)=F_\theta(x),\qquad
    (h_k,c_k)=F_\theta(h_{k-1},c_{k-1}),\quad k=1,\ldots,K,
    \label{eq:ar}
\end{equation}
where $F_\theta$ is the MLLM backbone, $h_k$ the hidden state at step $k$, and
$c_k$ the running KV cache. After $K$ transitions, a final embedding token reads
out the representation. Recurrence enables input-dependent refinement but
remains deterministic: each input follows a single fixed trajectory. VaME
instead learns a conditional distribution over recurrent trajectories, enabling
exploration during training while following the conditional means at inference.

%% file: sections/method.tex
\section{Variational Multimodal Embeddings}
\label{sec:method}

\begin{figure}[t]
    \centering
    \includegraphics[width=\linewidth]{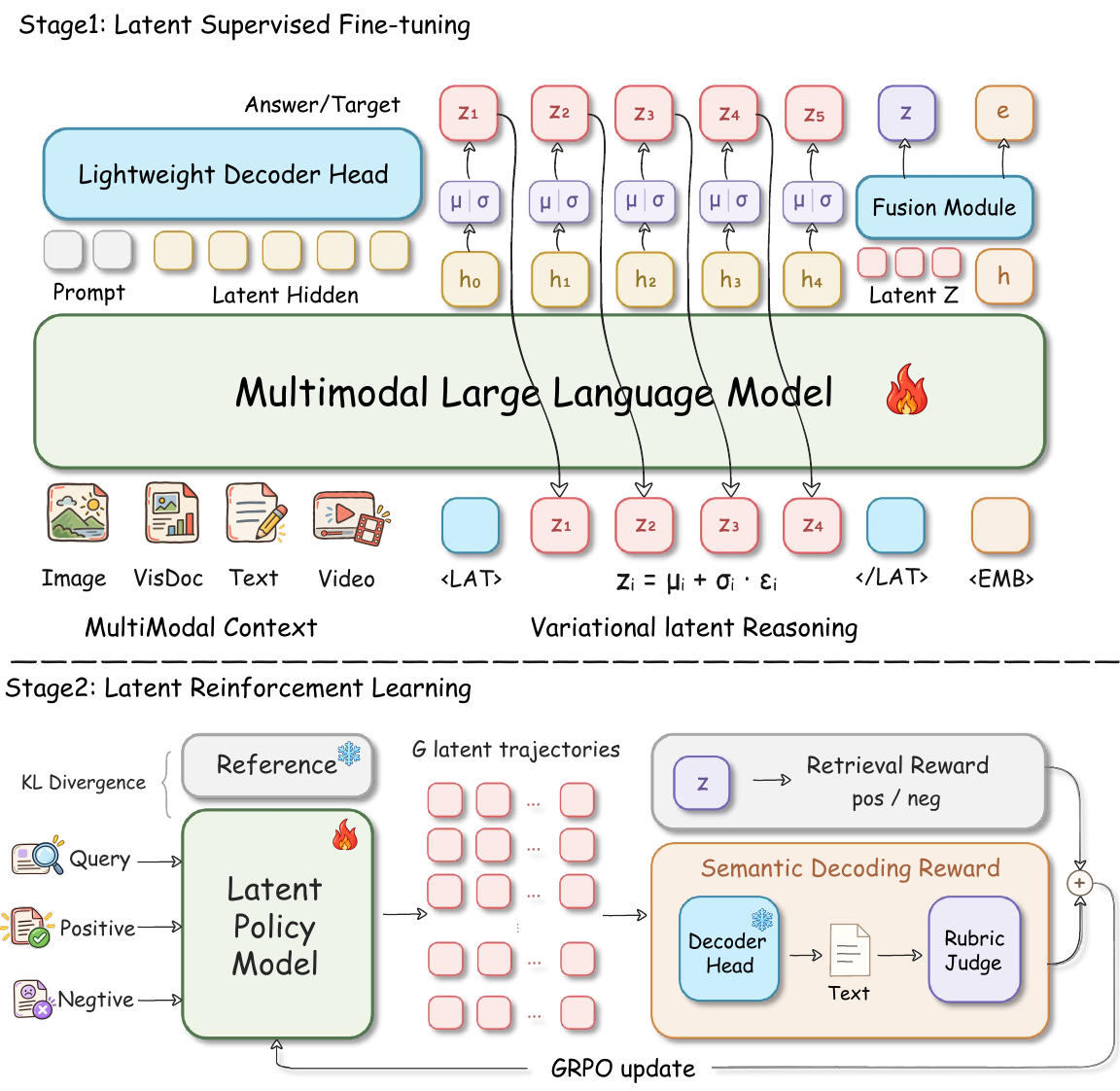}
    \caption{Overview of the \method{} framework. Latent SFT aligns variational
    trajectories with retrieval and answer supervision. Latent RL compares
    sampled trajectories and refines the policy using retrieval feedback and
    the semantic decoding reward.}
    \label{fig:framework}
\end{figure}

As shown in Figure~\ref{fig:framework}, \method{} establishes a two-stage training
framework for learning multimodal embeddings through latent computation. Its
central challenge is not merely to introduce noise or stochasticity, but to make
sampled trajectories
represent distinct computations, align their variation with retrieval, and
confine exploration to training. Accordingly, Section~\ref{sec:vae} formulates
autoregressive computation as a regularized conditional trajectory distribution;
Section~\ref{sec:supervision} aligns its samples with retrieval relevance and
answer semantics through latent SFT and complementary readouts;
Section~\ref{sec:sdr} compares alternative trajectories using retrieval feedback
and SDR, then improves the latent policy with group-relative RL. At inference,
the learned policy follows its conditional means, retaining deterministic
single-vector retrieval.


\subsection{Variational Latent Reasoning}
\label{sec:vae}

\noindent\textbf{Stochastic latent recurrence.}\ 
The latent recurrence in Equation~\ref{eq:ar} enables input-dependent
refinement, but maps each input to only one deterministic trajectory. This
leaves no alternative computations for exploration or outcome-based comparison.
We therefore introduce stochasticity into the recurrent transitions, allowing
multiple latent trajectories to be sampled from the same multimodal input. VLR
realizes this idea by replacing each deterministic transition with a conditional
diagonal Gaussian. As shown in Figure~\ref{fig:framework}, we introduce two special
tokens, \texttt{<LAT>} and \texttt{</LAT>}, to mark the start and end of latent
reasoning, respectively. Given an input $x$, a multimodal prefill produces an initial
cache $c_0$ and hidden state $h_0$. At step $k$, a variational head predicts the
distribution parameters from the current hidden state, draws noise
$\epsilon_k\sim\mathcal N(0,I)$ to construct a continuous latent action, and
feeds it back into the backbone:
\begin{equation}
\begin{gathered}
    (\mu_k,\log\sigma_k)=g_\phi(h_{k-1}),\qquad
    z_k=\mu_k+\sigma_k\odot\epsilon_k,\\
    (h_k,c_k)=F_\theta(z_k,c_{k-1}),\qquad
    q_{\boldsymbol{\theta}}(Z\mid x)
    =\prod_{k=1}^{K}q_{\boldsymbol{\theta}}(z_k\mid x,z_{<k}),
\end{gathered}
\label{eq:vae}
\end{equation}
where $\boldsymbol{\theta}=(\theta,\phi)$ comprises the backbone parameters $\theta$ and
variational-head parameters $\phi$, and $Z=(z_1,\ldots,z_K)$ is the sampled latent
trajectory. Since each $z_k$ is processed before the next distribution is produced,
early samples affect the cache, hidden states, and subsequent action distributions.
Thus, $q_{\boldsymbol{\theta}}(Z\mid x)$ defines a distribution over complete latent
computations rather than independent noise added to the final embedding.
Sampling from this distribution yields alternative latent paths for the same
multimodal input, providing candidate trajectories for latent RL. The
reparameterization in Equation~\ref{eq:vae} allows gradients from both training
objectives to pass through the sampled latent actions, enabling joint optimization
of the backbone and variational head.

\noindent\textbf{Latent transition regularization.}\ 
Without a constraint on the conditional distributions, the policy could encode
training instances through arbitrarily displaced means or variances. We anchor
each transition to an isotropic Gaussian using
\begin{equation}
    \mathcal L_{\mathrm{KL}}
    =\frac{1}{K}\sum_{k=1}^{K}
      \KL\!\left(q_{\boldsymbol{\theta}}(z_k\mid x,z_{<k})\,\Vert\,\mathcal N(0,I)\right).
\label{eq:kl_main}
\end{equation}
The KL term keeps the trajectory space smooth enough for sampling, while the
task losses below determine which variations preserve relevance and answer
semantics. We use the term \emph{variational} for this conditional latent
distribution, reparameterized sampling, and prior regularization; the overall
retrieval objective is not an evidence lower bound for reconstructing the raw
multimodal input. Appendix~\ref{app:sft_losses} gives the exact Gaussian
parameterization and closed-form KL.

\subsection{Latent Supervised Fine-Tuning}
\label{sec:supervision}

The variational distribution above provides alternative latent computations, but
does not by itself identify which trajectories support retrieval. To supervise
these trajectories for retrieval, we complete the $K$ latent transitions and let
the embedding token read from the resulting cache, yielding the hidden state
$h$. To balance exploration with retrieval stability, we use
complementary readouts: the original embedding-token readout preserves a stable
single-vector interface, while the latent-fused embedding provides a
trajectory-sensitive signal for subsequent reinforcement learning:
\begin{equation}
    e=\norm{h},
    \qquad
    z=\norm{F_\omega(h,Z)},
\label{eq:aggregation}
\end{equation}
where $F_\omega$ aggregates the trajectory $Z$ via attention pooling and residually
combines it with $h$, parameterized by $\omega$.
For a paired batch $\mathcal B=\{(x_i^q,x_i^t)\}_{i=1}^{B}$, their joint
retrieval objective is
\begin{equation}
    \mathcal L_{\mathrm{ret}}
    =\mathcal L_{\mathrm{NCE}}^{e}
     +\lambda_z\mathcal L_{\mathrm{NCE}}^{z}
     +\lambda_Z\mathcal L_{\mathrm{NCE}}^{Z},
\label{eq:similarities}
\end{equation}
where $e$ and $z$ identify the two readouts, while $Z$ denotes position-aligned
pairwise scoring over latent actions. The first term learns the deployed geometry;
the latter two supervise trajectory-dependent retrieval, with $z$
also scoring latent-RL rollouts. At inference, only $e$ is used to
build the single-vector retrieval index.
Meanwhile, to align the trajectories with retrieval-relevant semantics, a lightweight decoder $p_\psi$
reconstructs query-side answers and target-side explanations or descriptions
from the projected post-action states $H=[h_1,\ldots,h_K]$:
\begin{equation}
    \mathcal L_{\mathrm{ans}}
    =-\mathbb E_{(x,y),\,Z\sim q_{\boldsymbol{\theta}}(\cdot\mid x)}
       \log p_\psi\!\left(y\mid \mathcal I_{\mathrm{dec}},H\right),
\label{eq:decode_main}
\end{equation}
where $\mathcal I_{\mathrm{dec}}$ denotes the fixed decoding prompt.
$\mathcal L_{\mathrm{div}}$ penalizes pairwise cosine similarity between
posterior means to prevent latent collapse. Together with KL regularization,
the retrieval and reconstruction signals yield the complete SFT objective:
\begin{equation}
    \mathcal L_{\mathrm{SFT}}
    =\mathcal L_{\mathrm{ret}}
     +\lambda_{\mathrm{ans}}\mathcal L_{\mathrm{ans}}
     +\lambda_{\mathrm{KL}}\mathcal L_{\mathrm{KL}}
     +\lambda_{\mathrm{div}}\mathcal L_{\mathrm{div}},
\label{eq:sft}
\end{equation}
where $\lambda_z$, $\lambda_Z$, $\lambda_{\mathrm{ans}}$,
$\lambda_{\mathrm{KL}}$, and $\lambda_{\mathrm{div}}$ weight the corresponding
objectives; exact settings are provided in Appendix~\ref{app:configuration}.

\subsection{Latent Reinforcement Learning}
\label{sec:sdr}

As illustrated in Equation~\ref{eq:sft}, latent SFT turns the variational
trajectory distribution into an aligned policy whose samples are grounded in
retrieval relevance and answer semantics. However, it does not express a
preference among the alternative computations available to the same query.
Latent RL supplies this missing comparison by refining the SFT policy under
group-relative retrieval and semantic feedback. Specifically, given a
query $q$, we draw $G$ trajectories $Z_g\sim\pi_{\mathrm{old}}(Z\mid q)$,
$g=1,\ldots,G$, and assign each trajectory two complementary rewards:
\begin{equation}
    R(Z_g)=R_{\mathrm{ret}}+\lambda_{\mathrm{SDR}}R_{\mathrm{SDR}},
\label{eq:sdr}
\end{equation}
where $R_{\mathrm{ret}}$ and $R_{\mathrm{SDR}}$ denote the retrieval and
semantic decoding rewards, respectively; $\lambda_{\mathrm{SDR}}$ balances
them, and $R(Z_g)$ is the trajectory-level reward used by GRPO.
Specifically, inspired by the similarity-gap reward in
UME-R1~\citep{umer1}, we make a simple refinement to $R_{\mathrm{ret}}$ by
softly weighting harder negatives. We complement this retrieval signal with
SDR, which evaluates whether a sampled trajectory retains retrieval-relevant
semantics:
\begin{equation}
    R_{\mathrm{ret}}=s_g^+-\sum_{j\in\mathcal N}w_{g,j}s_{g,j}^-,
    \qquad
    R_{\mathrm{SDR}}=J\!\left(\mathcal I_{\mathrm{eval}},y,
    F_\psi(\mathcal I_{\mathrm{dec}},H_g)\right),
\label{eq:ret_reward}
\end{equation}
where $\mathcal N$ is the negative candidate set, $s_g^+$ and $s_{g,j}^-$ are
positive and negative similarities for trajectory $Z_g$, and $w_{g,j}$ is
computed by applying softmax to its negative similarities.
For SDR, the frozen decoder $F_\psi$ greedily generates an answer from $H_g$
under $\mathcal I_{\mathrm{dec}}$, and the rubric judge $J$ in
Figure~\ref{fig:framework} scores its agreement with reference $y$ under
the rubric-based evaluation instruction $\mathcal I_{\mathrm{eval}}$.
Together, the two rewards favor retrieval-discriminative and semantically
grounded trajectories. The rubric and scoring details are provided in
Appendix~\ref{app:sdr_rubric}.
Given these trajectory-level rewards, we first standardize them within each group as
$A_g=(R(Z_g)-\bar R)/(\operatorname{std}_{g=1}^{G}[R(Z_g)]+\epsilon)$
and then minimize the following clipped group-relative loss for the continuous Gaussian policy:
\begin{align}
    \mathcal L_{\mathrm{GRPO}}={}&-\mathbb E_{q\sim\mathcal D,\,\{Z_g\}\sim\pi_{\mathrm{old}}(\cdot\mid q)}\!\Biggl[
    \frac{1}{G\cdot K}\sum_{g=1}^{G}\sum_{k=1}^{K}
    \Biggl(\min\!\Biggl\{
      \frac{\pi_{\boldsymbol{\theta}}(z_{g,k})}
           {\pi_{\mathrm{old}}(z_{g,k})}A_g,\notag\\[2pt]
    &\qquad\operatorname{clip}\!\left(
      \frac{\pi_{\boldsymbol{\theta}}(z_{g,k})}
           {\pi_{\mathrm{old}}(z_{g,k})},
      1-\epsilon_-,1+\epsilon_+\right)A_g\Biggr\}
    -\beta\KL\!\left(\pi_{\boldsymbol{\theta}}\Vert\pi_{\mathrm{ref}}\right)
    \Biggr)\Biggr],
\label{eq:grpo_main}
\end{align}
where $\mathcal D$ is the training query distribution and $K$ is the number
of latent steps. The current and rollout policies are $\pi_{\boldsymbol{\theta}}$ and
$\pi_{\mathrm{old}}$, respectively. For the step-$k$ action $z_{g,k}$ of
trajectory $g$, we compute its joint Gaussian probability density under each
policy, conditioned on the same query $q$ and preceding actions $z_{g,<k}$,
and take the current-to-rollout density ratio. The KL penalty uses the same
conditioning, with $\beta$ controlling regularization toward the frozen
reference policy $\pi_{\mathrm{ref}}$. Thus, the rewards rank the
sampled computations, and GRPO reinforces better trajectories while
limiting policy drift.

%% file: sections/experiments.tex
\section{Experiments}
\label{sec:experiments}


\input{tables/main_mmeb_v2}

\subsection{Experimental Setup}

\noindent\textbf{Implementation.}\ 
Following prior works~\citep{vlm2vecv2,plume,umer1,lame}, we initialize
the 2B and 7B \method{} variants from Qwen2-VL-2B-Instruct and
Qwen2-VL-7B-Instruct, respectively. Latent SFT trains on approximately
1.5 million image, video, and visual-document retrieval examples for
3,200 steps with $K=8$ latent steps, temperature $\tau=0.02$, and effective
batch size 512. The backbone and newly initialized modules use learning
rates of $10^{-5}$ and $2\times10^{-5}$, respectively. Latent RL trains
for 500 steps on approximately 10K records sampled from a separate pool of
approximately 127K candidates, with effective batch size 64 and learning rate
$10^{-6}$. We use $G=8$ trajectories per query,
GRPO clipping bounds $[\epsilon_-,\epsilon_+]=[0.8,1.28]$,
per-dimension KL weight
$\beta=0.01$, and
semantic-reward weight $\lambda_{\mathrm{SDR}}=0.2$.
We use Qwen3.5-122B-A10B via API as the judge model for SDR.

\noindent\textbf{Evaluation.} We evaluate broad multimodal retrieval on MMEB-V2~\citep{vlm2vecv2}, which
has 36 image and 18 video datasets covering retrieval, classification,
moment retrieval, and visual QA, all evaluated with Hit@1. Its 24 VisDoc
datasets involve document images such as complex charts and paper figures
and use nDCG@5. To assess reasoning-intensive retrieval,
we also evaluate on MRMR~\citep{mrmr}, which comprises 11 expert-level tasks.
We report Hit@1 for Negation and nDCG@10 for the remaining tasks, following
the official MRMR protocol.

\noindent\textbf{Baselines.}\ 
We compare \method{} with representative MLLM-based multimodal embedders
spanning three paradigms: direct-embedding models such as VLM2Vec, GME, and
Ops-MM-Embed; explicit-reasoning models TWN and UME-R1; and
latent-reasoning models PLUME, TTE-Flash, and LaME. Most baselines follow the training and
evaluation protocols of VLM2Vec-V2~\citep{vlm2vecv2}, making their performance
broadly comparable. We note that Ops-MM-Embed, GME, and LamRA use additional
training sources, while UME-R1 and TWN uses CoT from external reasoners to
enhance its embeddings.

\subsection{Main Results}

\noindent\textbf{Results on MMEB-V2.}\ 
As shown in Table~\ref{tab:main_mmeb_v2}, \method{}-2B and \method{}-7B score 65.7 and
69.4 on the 78-task MMEB-V2 benchmark, leading their respective size groups.
At 2B, \method{} outperforms LaME and PLUME by 1.3 and 4.1 points; relative to
LaME, it gains 0.2 on Image, 1.8 on Video, and 2.3 on VisDoc. It also surpasses
UME-R1 by 6.0 points at 2B and the explicit-reasoning baselines TWN and UME-R1
by 0.7 and 4.9 points at 7B, respectively. Thus, variational latent reasoning can improve
multimodal embeddings across modalities without generated reasoning text.

\input{tables/main_mrmr}

\noindent\textbf{Results on MRMR.}\ 
As shown in Table~\ref{tab:main_mrmr}, the 7B variant of \method{} scores 51.4
on MRMR, a benchmark comprising 11 retrieval tasks that require reasoning. It
outperforms RIME by 1.2 points and LaME, the deterministic baseline for latent
reasoning, by 1.6 points. It ranks first on six subtasks, including Medicine and
Science in Knowledge and all four Theorem domains, and improves over LaME
on nine tasks, including gains of 3.3, 3.0, and 2.8 points on Humanities, Math,
and Physics. \method{} also exceeds the explicit reasoning model UME-R1 by 3.4
points overall. These broad gains demonstrate robust retrieval on tasks that
require reasoning while retaining a latent interface based on a single vector
without generated reasoning text.

\subsection{Ablation Study}
\label{sec:ablations}
To validate the effectiveness of variational latent reasoning, latent
supervision, and reward-guided refinement, we conduct ablations on MMEB-V2
using the Qwen2-VL-2B embedder.

\begin{wraptable}[23]{r}{0.55\textwidth}
\vspace*{-\dimexpr\intextsep+\abovecaptionskip\relax}
\centering
{\input{tables/ablation_core_components}}
\vspace{0.25\baselineskip}
{\input{tables/ablation_training_objectives}}
\end{wraptable}

\noindent\textbf{Ablation on Core Components.}\ 
As shown in Table~\ref{tab:ablation_core_components}, we isolate VLR, Latent
SFT, and Latent RL under the same 2B setting. The deterministic AR baseline
scores 63.8, while VLR alone drops to 63.1, showing that unguided exploration
introduces noise. Latent SFT raises AR and VLR to 64.6 and 64.9, respectively,
demonstrating that direct supervision of latent trajectories is essential for
effective exploration. Latent RL further reaches 65.7, 1.9 points above the
AR baseline, validating the two-stage design in which supervision first grounds
latent exploration and reward guidance then refines the resulting trajectories.

\noindent\textbf{Ablation on Training Objectives.}\ 
As shown in Table~\ref{tab:ablation_training_objectives}, we ablate each
training objective under the same 2B setting while retaining
$\mathcal L_{\mathrm{NCE}}^{e}$. Removing $\mathcal L_{\mathrm{KL}}$ causes the
largest drop, from 65.7 to 64.4, highlighting the importance of variational
regularization for smooth sampling. Removing $\mathcal L_{\mathrm{div}}$ and
$\mathcal L_{\mathrm{ans}}$ costs 1.0 and 0.9 points, consistent with preventing
latent collapse and anchoring retrieval semantics. Removing
$\mathcal L_{\mathrm{NCE}}^{z}$ or $\mathcal L_{\mathrm{NCE}}^{Z}$ costs 0.9 or
0.4 points, showing complementary supervision of the fused latent readout and
per-step latent actions. All removals reduce the three modality averages,
indicating that each objective contributes to overall performance.

\input{tables/ablation_latent_rl_design}
\noindent\textbf{Ablation on Latent RL Rewards.}\ 
As shown in Table~\ref{tab:ablation_latent_rl_design}, we ablate retrieval
reward and SDR under a matched 2B setup. Latent SFT scores 64.9 and 38.3
on MMEB-V2 and MRMR, while retrieval reward
raises these scores to 65.2 and 38.5, while SDR reaches 65.3
and 40.7. Combining both rewards improves Image, Video, and VisDoc by 0.5,
0.8, and 1.1 points, respectively, and yields the best MMEB-V2 score of 65.7.
Its MRMR score reaches 40.5, 2.2 points above Latent SFT but 0.2 below SDR
alone. Overall, SDR substantially improves reasoning-intensive retrieval on MRMR.

%% file: tables/main_mmeb_v2.tex
\begin{table*}[!t]
\caption{Results on the MMEB-V2 benchmark. Bold and underline indicate the best
and second-best scores within each size group, respectively. CLS: classification, QA:
question answer, RET: retrieval, GD: grounding, MRET: moment retrieval,
VDR: ViDoRe, VR: VisRAG, OOD: out-of-distribution. Reported metrics adhere
to the settings of VLM2Vec-V2.}
\vspace{\baselineskip}
\centering
\resizebox{\textwidth}{!}{%
\begin{tabular}{l ccccc ccccc ccccc c}
\toprule
\multirow{2}{*}{\textbf{Model}}
& \multicolumn{5}{c}{\textbf{Image}}
& \multicolumn{5}{c}{\textbf{Video}}
& \multicolumn{5}{c}{\textbf{VisDoc}}
& \multirow{2}{*}{\textbf{All}} \\
\cmidrule(lr){2-6} \cmidrule(lr){7-11} \cmidrule(lr){12-16}
& \textbf{CLS} & \textbf{QA} & \textbf{RET} & \textbf{GD} & \textbf{Avg.}
& \textbf{CLS} & \textbf{QA} & \textbf{RET} & \textbf{MRET} & \textbf{Avg.}
& \textbf{VDRv1} & \textbf{VDRv2} & \textbf{VR} & \textbf{OOD} & \textbf{Avg.} & \\
\midrule
\textbf{\# of Datasets}
& 10 & 10 & 12 & 4 & 36
& 5 & 5 & 5 & 3 & 18
& 10 & 4 & 6 & 4 & 24 & 78 \\
\midrule
\rowcolor[HTML]{EDEDED}
\multicolumn{17}{c}{\emph{$\sim$2B Model Size}} \\
VLM2Vec
& 58.7 & 49.3 & 65.0 & 72.9 & 59.7
& 33.4 & 30.5 & 20.6 & 33.0 & 29.0
& 49.8 & 13.5 & 51.8 & 33.5 & 41.6 & 47.0 \\
DUME
& 59.3 & 55.0 & 66.3 & 78.0 & 62.5
& 37.7 & 46.6 & 17.1 & 30.0 & 33.2
& 67.6 & 43.3 & 47.1 & 33.8 & 52.8 & 52.7 \\
GME
& 54.4 & 29.9 & 66.9 & 55.5 & 51.9
& 34.9 & 42.0 & 25.6 & 32.4 & 33.9
& \textbf{86.1} & \underline{54.0} & 82.5 & 43.1 & \underline{72.7} & 54.1 \\
VLM2Vec-V2
& 62.0 & 56.3 & 69.5 & 77.3 & 64.9
& 39.3 & 34.3 & 28.8 & 38.5 & 34.9
& 75.5 & 44.9 & 79.4 & 39.4 & 65.4 & 58.0 \\
UME-R1
& 64.8 & 62.8 & 67.6 & 77.2 & 66.6
& 44.3 & 51.0 & 32.9 & 39.7 & 42.2
& 72.4 & 46.2 & 79.2 & 28.9 & 62.5 & 59.7 \\
PLUME
& 66.5 & 59.2 & 67.6 & 79.7 & 66.3
& 45.0 & 52.3 & 33.5 & \underline{46.7} & 44.1
& 72.1 & 49.8 & 78.1 & \underline{57.4} & 67.5 & 61.6 \\
Ops-MM-Embed
& \textbf{68.1} & 65.1 & 69.2 & 80.9 & 69.0
& \underline{53.6} & \textbf{55.7} & \underline{41.8} & 33.7 & \underline{47.6}
& 76.4 & 53.2 & 77.6 & 32.6 & 65.5 & 63.0 \\
TTE-Flash
& 67.1 & 61.7 & \textbf{70.7} & \textbf{81.3} & 68.3
& \textbf{54.4} & 51.5 & \textbf{45.4} & \textbf{51.5} & \textbf{50.6}
& 75.6 & 52.7 & \underline{83.8} & 41.1 & 68.1 & 64.1 \\
LaME
& \underline{67.6} & \underline{66.2} & \underline{70.5} & \underline{81.2} & \underline{69.3}
& 45.6 & 52.3 & 36.3 & 43.5 & 44.5
& 78.2 & 48.4 & 82.8 & 55.4 & 72.1 & \underline{64.4} \\
\rowcolor{papertint}
\textbf{\method{} (Ours)}
& \underline{67.6} & \textbf{66.6} & 70.0 & 80.3 & \textbf{69.5}
& 46.2 & \underline{54.8} & 38.2 & 45.5 & 46.3
& \underline{79.4} & \textbf{55.2} & \textbf{85.6} & \textbf{64.1} & \textbf{74.4} & \textbf{65.7} \\
\midrule
\rowcolor[HTML]{EDEDED}
\multicolumn{17}{c}{\emph{7--8B Model Size}} \\
LamRA
& 51.7 & 34.1 & 66.9 & 56.7 & 52.4
& 32.9 & 42.6 & 23.2 & 37.6 & 33.7
& 56.3 & 33.3 & 58.2 & 40.1 & 50.2 & 47.4 \\
VLM2Vec
& 62.7 & 56.9 & 69.4 & 82.2 & 65.5
& 39.1 & 30.0 & 29.0 & 40.6 & 34.0
& 56.9 & 9.4 & 59.1 & 38.1 & 46.4 & 52.3 \\
DUME
& 64.2 & 57.0 & 70.8 & 81.8 & 66.4
& 32.9 & 47.4 & 8.6 & 28.0 & 29.4
& 67.1 & 35.2 & 82.6 & 34.9 & 60.3 & 55.9 \\
GME
& 57.7 & 34.7 & 71.2 & 59.3 & 56.0
& 37.4 & 50.4 & 28.4 & 38.2 & 38.6
& \textbf{89.4} & 55.6 & 85.0 & 44.4 & 75.2 & 57.8 \\
CAFe
& 63.6 & 61.7 & 69.1 & \textbf{87.6} & 67.6
& 35.8 & 58.7 & 34.4 & 39.5 & 42.4
& 70.7 & 49.6 & 79.5 & 38.1 & 63.9 & 60.6 \\
UME-R1
& 67.1 & 69.2 & 71.9 & 84.9 & 71.3
& 48.6 & 60.7 & 38.2 & 39.3 & 47.5
& 75.7 & 50.5 & 83.7 & 37.6 & 67.1 & 64.5 \\
TWN
& \underline{70.2} & \textbf{74.3} & 70.8 & 87.1 & \textbf{73.4}
& 50.1 & \underline{64.0} & 34.8 & 40.8 & 48.2
& 82.5 & 56.7 & \underline{86.2} & \textbf{70.0} & \underline{77.0} & 68.7 \\
LaME
& 70.0 & \underline{71.0} & \underline{73.1} & 85.8 & 73.0
& \underline{51.0} & \textbf{64.8} & 40.3 & \underline{44.6} & \underline{50.8}
& 81.5 & 58.2 & 85.0 & \underline{68.2} & 75.9 & 68.8 \\
Ops-MM-Embed
& 69.7 & 69.6 & \underline{73.1} & \underline{87.2} & 72.7
& \textbf{59.7} & 62.2 & \textbf{45.7} & 43.2 & \textbf{53.8}
& 80.1 & \textbf{59.6} & 79.3 & 67.8 & 74.4 & \underline{68.9} \\
\rowcolor{papertint}
\textbf{\method{} (Ours)}
& \textbf{71.3} & 70.3 & \textbf{73.7} & 84.1 & \underline{73.2}
& 49.9 & 58.6 & \underline{42.0} & \textbf{49.3} & 50.0
& \underline{83.2} & \underline{59.0} & \textbf{89.4} & 67.9 & \textbf{78.2} & \textbf{69.4} \\
\bottomrule
\end{tabular}%
}
\label{tab:main_mmeb_v2}
\end{table*}

%% file: tables/main_mrmr.tex
\begin{table*}[!t]
\caption{Results on the reasoning-intensive MRMR benchmark including Art, Medicine (Med.), Science (Sci.), Humanities (Hum.), Math, Physics (Phy.), Engineering (Eng.), Business (Bus.), Negation (Neg.), Design (Des.), and Traffic (Tra.). Bold and underlined scores denote the best and second-best performance, respectively.}
\label{tab:main_mrmr}
\begin{center}
\scriptsize
\setlength{\tabcolsep}{4.5pt}
\renewcommand{\arraystretch}{1.05}
\resizebox{\textwidth}{!}{%
\begin{tabular}{ll cccc cccc ccc c}
\toprule
\multirow{2}{*}{\textbf{Model}} 
& \multirow{2}{*}{\textbf{Backbone}}
& \multicolumn{4}{c}{\textbf{Knowledge}} 
& \multicolumn{4}{c}{\textbf{Theorem}} 
& \multicolumn{3}{c}{\textbf{Contradiction}} 
& \multirow{2}{*}{\textbf{All}} \\
\cmidrule(lr){3-6} \cmidrule(lr){7-10} \cmidrule(lr){11-13}
& & Art & Med. & Sci. & Hum. & Math & Phy. & Eng. & Bus. & Neg. & Des. & Tra. & \\
\midrule
E5-V & LLaVA-Next-8B & 25.1 & 11.7 & 16.6 & 10.8 & 2.1 & 3.4 & 2.5 & 5.2 & \underline{11.5} & 3.7 & 2.1 & 8.6 \\
EVA-CLIP & EVA-ViT-0.4B & 10.2 & 13.5 & 26.1 & 12.9 & 6.2 & 10.5 & 9.3 & 11.7 & 8.5 & 4.4 & 5.4 & 10.8 \\
OpenCLIP & ViT-G/14-1B & 56.0 & 17.9 & 33.2 & 22.0 & 5.7 & 5.0 & 7.0 & 9.7 & 13.0 & 8.1 & 12.4 & 17.3 \\
VLM2Vec & Qwen2-VL-7B & 53.5 & 22.4 & 36.7 & 24.0 & 2.1 & 2.8 & 2.8 & 2.9 & \underline{11.5} & 5.6 & 18.3 & 18.1 \\
VISTA & Qwen2-VL-2B & 21.3 & 27.8 & 32.6 & 17.0 & 18.8 & 17.1 & 17.3 & 28.6 & 20.0 & 20.2 & 9.4 & 20.9 \\
ColPali & PaliGemma-3B & 36.1 & 29.9 & 42.7 & 29.2 & 5.7 & 14.8 & 12.0 & 24.6 & \textbf{28.5} & 19.4 & 18.2 & 23.7 \\
GME & Qwen2-VL-7B & 54.3 & 40.1 & 46.8 & 45.6 & 28.8 & 36.0 & 30.2 & 45.1 & 15.0 & 26.3 & 29.6 & 36.2 \\
MM-Embed & NV-Embed-8B & 65.6 & 53.0 & 63.5 & 62.8 & 23.6 & 30.8 & 27.4 & 44.9 & 7.0 & 23.8 & 34.9 & 39.8 \\
UME-R1 & Qwen2-VL-7B & \underline{77.8} & 55.7 & 72.9 & 64.1 & 27.2 & 39.2 & 32.2 & 47.8 & 7.5 & 61.9 & \underline{41.7} & 48.0 \\
Ops-MM-Embed & Qwen2-VL-7B & \textbf{79.3} & 52.5 & 70.0 & 67.8 & 27.7 & 39.5 & 30.1 & \underline{52.3} & 8.0 & 55.9 & \textbf{45.8} & 48.1 \\
LaME & Qwen2-VL-7B & 73.4 & 58.2 & \underline{73.8} & 65.6 & \underline{29.5} & \underline{44.4} & \underline{36.4} & 52.2 & 8.5 & \textbf{64.9} & 40.9 & 49.8 \\
RIME & Qwen2-VL-7B & 76.8 & \underline{58.5} & 73.6 & \textbf{71.4} & 29.2 & 43.0 & 35.6 & 51.8 & 8.5 & \underline{64.1} & 39.5 & \underline{50.2} \\
\rowcolor{papertint}
\textbf{\method{} (Ours)} & Qwen2-VL-7B & 75.6 & \textbf{60.7} & \textbf{76.3} & \underline{68.9} & \textbf{32.5} & \textbf{47.2} & \textbf{37.8} & \textbf{52.9} & 10.1 & 63.7 & 39.7 & \textbf{51.4} \\
\bottomrule
\end{tabular}%
}
\end{center}
\end{table*}

%% file: tables/ablation_core_components.tex
\caption{Ablation on core components on MMEB-V2. All crosses
($\times\,\times\,\times$) denote the deterministic AR baseline.
All reports the overall MMEB-V2 score.}
\vspace{0.5\baselineskip}
\label{tab:ablation_core_components}
\small
\setlength{\tabcolsep}{2.5pt}
\renewcommand{\arraystretch}{0.9}
\resizebox{\linewidth}{!}{%
\begin{tabular}{ccccccc}
\toprule
VLR & Latent SFT & Latent RL & Image & Video & VisDoc & All \\
\midrule
$\times$ & $\times$ & $\times$
& 68.5 & 43.9 & 71.3 & 63.8 \\
$\times$ & $\checkmark$ & $\times$
& 68.9 & 45.1 & 72.8 & 64.6 \\
$\checkmark$ & $\times$ & $\times$
& 67.8 & 43.3 & 70.9 & 63.1 \\
$\checkmark$ & $\checkmark$ & $\times$
& 69.0 & 45.5 & 73.3 & 64.9 \\
\rowcolor{papertint}
$\checkmark$ & $\checkmark$ & $\checkmark$
& 69.5 & 46.3 & 74.4 & 65.7 \\
\bottomrule
\end{tabular}
}

%% file: tables/ablation_training_objectives.tex
\caption{Ablation on training objectives on MMEB-V2. VaME-2B uses the full
objective in Equation~\ref{eq:sft}. All reports the overall MMEB-V2 score.}
\vspace{0.5\baselineskip}
\label{tab:ablation_training_objectives}
\setlength{\tabcolsep}{10pt}
\renewcommand{\arraystretch}{1.01}
\resizebox{\linewidth}{!}{%
\begin{tabular}{clcccc}
\toprule
\# & Model & Image & Video & VisDoc & All \\
\midrule
\rowcolor{papertint}
1 & VaME-2B & 69.5 & 46.3 & 74.4 & 65.7 \\
\midrule
2 & w/o $\mathcal L_{\mathrm{NCE}}^{z}$
& 68.8 & 45.1 & 73.6 & 64.8 \\
3 & w/o $\mathcal L_{\mathrm{NCE}}^{Z}$
& 69.3 & 46.2 & 73.6 & 65.3 \\
4 & w/o $\mathcal L_{\mathrm{ans}}$
  & 68.8 & 45.7 & 73.1 & 64.8 \\
5 & w/o $\mathcal L_{\mathrm{KL}}$
  & 68.4 & 45.2 & 72.7 & 64.4 \\
6 & w/o $\mathcal L_{\mathrm{div}}$
  & 68.7 & 45.5 & 73.1 & 64.7 \\

\bottomrule
\end{tabular}
}

%% file: tables/ablation_latent_rl_design.tex
\begin{wraptable}{r}{0.55\textwidth}
\vspace*{-\dimexpr\intextsep+\abovecaptionskip\relax}
\vspace*{\baselineskip}
\centering
\caption{Ablation on latent RL rewards on MMEB-V2. Latent RL combines the
retrieval reward and SDR. All reports the overall MMEB-V2 score.}
\vspace{0.5\baselineskip}
\label{tab:ablation_latent_rl_design}
\footnotesize
\setlength{\tabcolsep}{0.96\tabcolsep}
\renewcommand{\arraystretch}{1.08}
\resizebox{\linewidth}{!}{%
\begin{tabular}{clccccc}
\toprule
\# & Model & Image & Video & VisDoc & All & MRMR \\
\midrule
1 & Latent SFT & 69.0 & 45.5 & 73.3 & 64.9 & 38.3 \\
2 & w/ ret reward & 69.0 & 45.8 & 73.8 & 65.2 & 38.5 \\
3 & w/ SDR & 69.1 & 46.2 & 74.1 & 65.3 & 40.7 \\
\rowcolor{papertint}
4 & VaME-2B & 69.5 & 46.3 & 74.4 & 65.7 & 40.5 \\
\bottomrule
\end{tabular}
}
\end{wraptable}

%% file: sections/deep_analysis.tex
\section{Deep Analysis}

\begin{wrapfigure}{l}{0.62\textwidth}
\vspace*{-\intextsep}
\centering
\includegraphics[width=\linewidth]{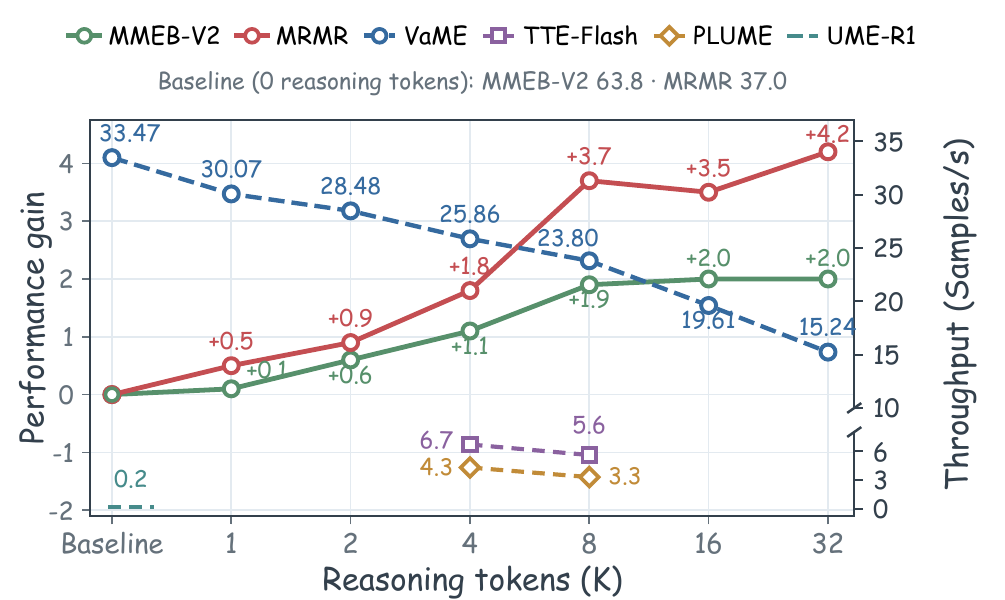}
\setlength{\abovecaptionskip}{-0.5\baselineskip}
\caption{Effect of latent tokens on gains (solid lines) and throughput (dashed lines).}
\label{fig:ablationk}
\vspace{0.25\baselineskip}
\includegraphics[width=\linewidth]{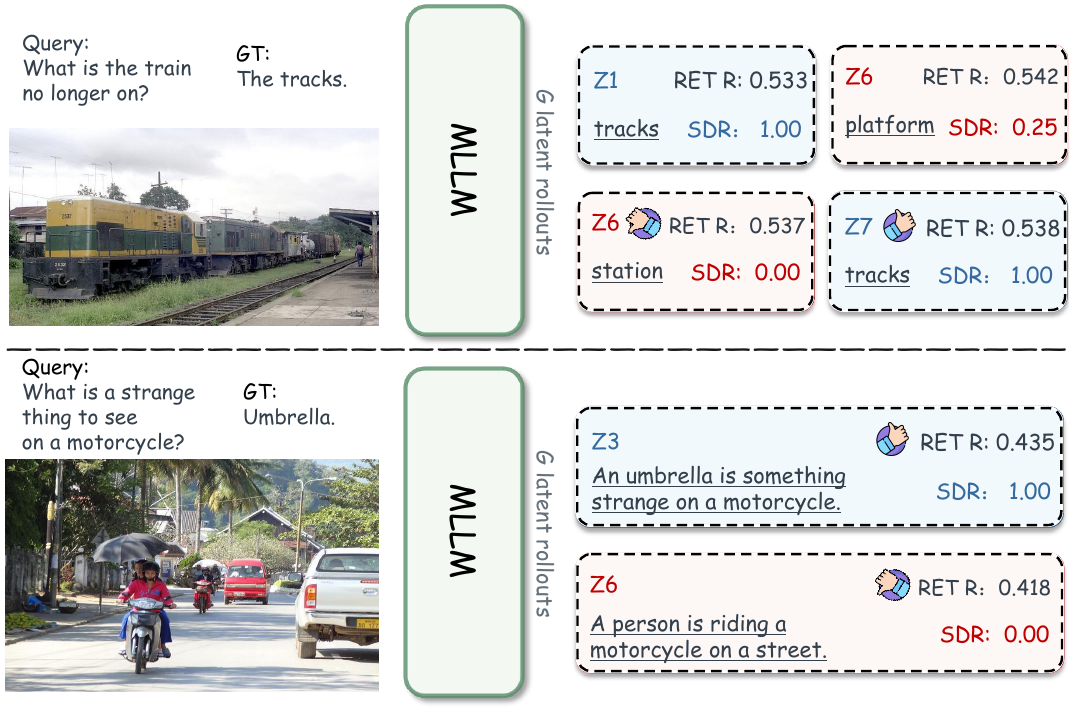}
\setlength{\abovecaptionskip}{-0.5\baselineskip}
\caption{Qualitative latent-rollout visualization.}
\label{fig:latent_rollout_visualization}
\vspace*{-0.5\intextsep}
\end{wrapfigure}

\textbf{Analysis of Latent Token Budget.}\ 
Figure~\ref{fig:ablationk} compares performance and throughput across latent token
budgets, with throughput measured on a single H800 GPU at batch size 1.
At $K=8$, VaME improves MMEB-V2 and MRMR by 1.9 and 3.7 points over the baseline,
respectively, while achieving at least a $4.25\times$ speedup over TTE-Flash and
PLUME at the same budget.
Larger budgets yield limited additional gains while reducing throughput,
supporting $K=8$ for the main experiments.

\vspace{-0.36\baselineskip}
\textbf{Analysis of Trajectory Diversity.}\ 
As shown in Figure~\ref{fig:latent_rollout_visualization}, SDR rewards recovering the
queried fact. ``Tracks'' answers what the train is no longer on; ``platform'' is only
contextually related, while ``station'' gives the wrong location. Likewise,
``umbrella'' identifies the unusual object omitted by the street description.
Yet ``platform'' earns higher retrieval reward than both correct ``tracks'' trajectories.
Retrieval reward measures embedding separation rather than answer fidelity,
explaining this mismatch and motivating SDR as complementary semantic supervision.

\WFclear

%% file: sections/conclusion.tex
\section{Conclusion}

We present \method{}, a variational multimodal embedding framework that models
latent reasoning as autoregressive trajectory distributions. Latent SFT grounds
trajectories in retrieval relevance and answer semantics, while Latent RL
refines them with retrieval feedback and SDR. At inference, one conditional-mean
trajectory yields a single embedding without textual reasoning or search.
Results on MMEB-V2 and MRMR show that \method{} outperforms prior methods in
both broad and reasoning-intensive retrieval with substantially faster
inference. Future work will explore adaptive computation to further reduce
overhead and align stochastic training with deterministic inference.

%% file: sections/statements.tex
\subsection*{AI use statement}
An AI coding assistant was used to draft and revise the manuscript, inspect
the local implementation and evaluation summaries, generate vector figures
and tables, and check the LaTeX build. An additional AI agent was used to
review the draft. No missing experimental results were generated or imputed.
Human author review and verification are required before submission.

%% file: sections/appendix.tex
\section{Training Data and Setup}
\label{app:data}

\begin{wraptable}{r}{0.64\textwidth}
    \vspace{-\dimexpr\intextsep+\baselineskip\relax}
    \centering
    \setlength{\tabcolsep}{5pt}
    \small
    \caption{Statistics of the latent-SFT training data composition.}
    \label{tab:appendix_data}
    \vspace{0.5\baselineskip}
    \resizebox{\linewidth}{!}{%
        \begin{tabular}{@{}lrrrl@{}}
            \toprule
            \textbf{Dataset} & \textbf{Initial} & \textbf{Final SFT} &
            \textbf{Ratio} & \textbf{Retrieval direction} \\
            \midrule
            \rowcolor{papertint}
            \multicolumn{5}{c}{\textit{Image-based (MMEB-train)}} \\
            A-OKVQA & 37,929 & 22,685 & 59.81\% & Image--text $\rightarrow$ text \\
            CIRR & 35,085 & 32,203 & 91.79\% & Image--text $\rightarrow$ image--text \\
            ChartQA & 39,512 & 34,962 & 88.48\% & Image--text $\rightarrow$ text \\
            DocVQA & 47,401 & 42,700 & 90.08\% & Image--text $\rightarrow$ text \\
            HatefulMemes & 16,572 & 10,612 & 64.04\% & Image--text $\rightarrow$ text \\
            ImageNet-1K & 44,409 & 40,940 & 92.19\% & Image--text $\rightarrow$ text \\
            InfographicsVQA & 40,746 & 34,119 & 83.74\% & Image--text $\rightarrow$ text \\
            MSCOCO & 26,429 & 19,111 & 72.31\% & Image--text $\rightarrow$ image--text \\
            MSCOCO-i2t & 46,596 & 42,723 & 91.69\% & Image--text $\rightarrow$ text \\
            MSCOCO-t2i & 43,173 & 39,698 & 91.95\% & Text $\rightarrow$ image--text \\
            N24News & 30,320 & 26,072 & 85.99\% & Image--text $\rightarrow$ text \\
            NIGHTS & 43,167 & 25,847 & 59.88\% & Image--text $\rightarrow$ image--text \\
            OK-VQA & 19,900 & 12,641 & 63.52\% & Image--text $\rightarrow$ text \\
            SUN397 & 45,864 & 33,373 & 72.77\% & Image--text $\rightarrow$ text \\
            VOC2007 & 20,454 & 11,952 & 58.43\% & Image--text $\rightarrow$ text \\
            Visual7W & 41,677 & 38,479 & 92.33\% & Image--text $\rightarrow$ text \\
            VisDial & 34,652 & 31,915 & 92.10\% & Text $\rightarrow$ image--text \\
            VisualNews-i2t & 34,364 & 31,568 & 91.86\% & Image--text $\rightarrow$ text \\
            VisualNews-t2i & 28,684 & 26,383 & 91.98\% & Text $\rightarrow$ image--text \\
            WebQA & 43,910 & 26,981 & 61.45\% & Text $\rightarrow$ image--text \\[2pt]

            \rowcolor{papertint}
            \multicolumn{5}{c}{\textit{Video-based (LLaVA-Hound)}} \\
            Caption Retrieval & 284,721 & 261,925 & 91.99\% & Video $\rightarrow$ text \\
            Video QA & 274,006 & 252,242 & 92.06\% & Video--text $\rightarrow$ text \\
            Video Retrieval & 260,410 & 239,462 & 91.96\% & Text $\rightarrow$ video \\[2pt]

            \rowcolor{papertint}
            \multicolumn{5}{c}{\textit{Document-based}} \\
            ViDoRe & 83,964 & 77,265 & 92.02\% & Image--text $\rightarrow$ text \\
            VisRAG & 60,266 & 55,462 & 92.03\% & Text $\rightarrow$ image \\
            \midrule
            \textbf{Image-based} & 720,844 & 584,964 & 81.15\% & Image-centric \\
            \textbf{Video-based} & 819,137 & 753,629 & 92.00\% & Video-centric \\
            \textbf{Document-based} & 144,230 & 132,727 & 92.02\% & Document-centric \\
            \textbf{Total} & \textbf{1,684,211} & \textbf{1,471,320} &
            \textbf{87.36\%} & Multimodal \\
            \bottomrule
        \end{tabular}%
    }
    \vspace*{-0.5\intextsep}
\end{wraptable}

Latent SFT retains the public multimodal mixture used by the LaME training
corpus~\citep{lame}, itself derived from the VLM2Vec-V2 collection~\citep{vlm2vecv2}.
Table~\ref{tab:appendix_data} reports the modality-level composition rather than
only the aggregate size. The mixture covers image--text matching,
classification, visual question answering, composed retrieval, video QA and
retrieval, and document-image retrieval. We reuse existing answer, description,
or explanation fields as decoder targets; no intermediate chain-of-thought
annotation is required.
The initial count refers to records in the merged source pool. The final SFT
count is measured after normalized-pair deduplication and an entity-level split
that reserves approximately $8\%$ of unique pairs as unseen RL candidates.

Latent RL draws from a separately curated pool of approximately 127K multimodal
candidates. However, the reported RL run uses approximately 10K records sampled from this
pool for policy optimization. Each selected record supplies a query, its paired
positive, candidate targets, and a reference answer for semantic scoring. The
candidate pool is used only for refinement; MMEB-V2 and MRMR remain held-out
evaluation benchmarks.

\section{Variational Latent Learning}
\label{app:implementation}

Building on Section~\ref{sec:method}, this section details the dependency between
the two optimization stages. Latent SFT first grounds the stochastic trajectory
distribution in retrieval relevance and answer semantics; initialized from this
aligned policy, latent RL then uses outcome feedback to favor more discriminative
and semantically faithful trajectories.

\subsection{Latent SFT Loss Formulation}
\label{app:sft_losses}

As shown in Table~\ref{tab:objective_roles}, the latent-SFT objective comprises
five groups of losses with complementary roles. The three retrieval losses share
the contrastive objective but supervise different representations:
$\mathcal L_{\mathrm{NCE}}^{e}$ shapes the deployed embedding-token geometry,
$\mathcal L_{\mathrm{NCE}}^{z}$ makes complete rollouts comparable, and
$\mathcal L_{\mathrm{NCE}}^{Z}$ directly supervises the step-wise latent actions.
$\mathcal L_{\mathrm{ans}}$ preserves answer-level semantics, while
$\mathcal L_{\mathrm{KL}}$ and $\mathcal L_{\mathrm{div}}$ regularize the
conditional Gaussian trajectory. The main text already defines the embedding-token
and fused-rollout objectives. Below, we give the additional implementation details
for the position-aligned latent loss, answer reconstruction loss, and Gaussian
regularization used throughout latent optimization.

For $\mathcal L_{\mathrm{NCE}}^{Z}$, the position-aligned similarity between
query and target is
\begin{equation}
    s^Z(q,t)
    =\frac{1}{K}\sum_{k=1}^{K}
      \left\langle \hat z^q_k,\hat z^t_k\right\rangle,
    \label{eq:position_aligned_similarity}
\end{equation}
where $\hat z_k=\norm{z_k}$ denotes the normalized latent action at step $k$.

\begin{center}
    \refstepcounter{table}\label{tab:objective_roles}
    \begin{minipage}{\textwidth}
    \centering
    \small
    Table~\thetable: Functional role of the latent-SFT objectives.\\[4pt]
    \setlength{\tabcolsep}{6pt}
    \renewcommand{\arraystretch}{1.12}
    \begin{tabular*}{\linewidth}{@{}
        >{\raggedright\arraybackslash}p{0.17\linewidth}
        @{\extracolsep{\fill}}
        >{\raggedright\arraybackslash}p{0.27\linewidth}
        >{\raggedright\arraybackslash}p{0.50\linewidth}@{}}
        \toprule
        Objective & Directly supervised object & Training role \\
        \midrule
        $\mathcal L_{\mathrm{NCE}}^{e}$
            & embedding-token readout $e$
            & learns the single-vector query--target geometry used by the index \\
        $\mathcal L_{\mathrm{NCE}}^{z}$
            & fused rollout readout $z$
            & makes complete sampled trajectories comparable for latent RL \\
        $\mathcal L_{\mathrm{NCE}}^{Z}$
            & position-aligned actions $z_{1:K}$
            & supplies retrieval supervision throughout the recurrent trajectory \\
        $\mathcal L_{\mathrm{ans}}$
            & post-action states $h_{1:K}$
            & anchors latent computation to answer-level semantics \\
        $\mathcal L_{\mathrm{KL}}+\mathcal L_{\mathrm{div}}$
            & Gaussian parameters $\mu_{1:K},\sigma_{1:K}$
            & controls sampling and discourages step-wise latent collapse \\
        \bottomrule
    \end{tabular*}
    \end{minipage}
\end{center}

$\mathcal L_{\mathrm{NCE}}^{Z}$ applies Equation~\ref{eq:contrastive} using
the pairwise scores $s^Z(q,t)$. Unlike $\mathcal L_{\mathrm{NCE}}^{z}$, which supervises the fused
rollout readout, this term supplies a retrieval signal directly to every
position in the latent trajectory. All three retrieval losses use temperature
$\tau=0.02$ and the same masked, hard-negative-weighted implementation.

For $\mathcal L_{\mathrm{ans}}$, a decoder $p_\psi$ conditions on the fixed
prompt $\mathcal I_{\mathrm{dec}}$ and post-action states $H$:
\begin{equation}
    \mathcal L_{\mathrm{ans}}^{(s)}
    =-\frac{1}{N_s}\sum_{i\in\mathcal B_s}\sum_{t\in\mathcal V_i}
       \log p_\psi(y^i_t\mid y^i_{<t},\mathcal I_{\mathrm{dec}},H_i),
    \label{eq:decode}
\end{equation}
where $\mathcal B_s$ is the batch for logical side $s$, $\mathcal V_i$ contains
valid answer-token positions, and $N_s=\sum_i|\mathcal V_i|$. The loss is
averaged across supervised sides. Padding and prefix positions are excluded;
retained end tokens are included. Training annotations provide the answer
targets, not intermediate chain-of-thought text.
The specific form of the decoder prompt $\mathcal I_{\mathrm{dec}}$ is shown in
Figure~\ref{fig:decoder_interface}.

\begin{figure}[h!]
    \centering
    \setlength{\fboxsep}{5pt}
    \fcolorbox{paperblue}{papertint}{%
        \begin{minipage}{0.91\columnwidth}
        \small
        \textcolor{paperblue}{\textbf{Fixed instruction}}\quad
        Decode the following latent representations into the answer.\\[3pt]
        \textcolor{paperteal}{\textbf{Continuous prefix}}\quad
        $P(h_1)\;P(h_2)\;\cdots\;P(h_K)$\\[3pt]
        \textcolor{paperblue}{\textbf{Assistant target}}\quad
        $y_1\;y_2\;\cdots\;y_T\;\langle\mathrm{eos}\rangle$
        \end{minipage}}
    \caption{Illustration of the decoder prompt.}
    \label{fig:decoder_interface}
\end{figure}

For $\mathcal L_{\mathrm{KL}}$, two learned projections parameterize the
diagonal Gaussian at latent step $k$, and the closed-form standard-normal
regularizer is
\begin{equation}
\begin{aligned}
    \mathcal L_{\mathrm{KL}}
    &=\frac{1}{2Kd}\sum_{k=1}^{K}\sum_{j=1}^{d}
      \left(\mu_{kj}^{2}+\sigma_{kj}^{2}-1-2\log\sigma_{kj}\right).
\end{aligned}
    \label{eq:kl}
\end{equation}
where $K$ is the number of latent steps and $d$ is the latent-state dimension.
The KL term is averaged over examples and both retrieval sides. Its relation to
the conditional answer-likelihood bound is derived in Appendix~\ref{app:elbo}.
$\mathcal L_{\mathrm{div}}$ is the mean
absolute off-diagonal cosine similarity among posterior means and discourages
step-wise latent collapse.

\subsection{Latent RL Training Details}
\label{app:sdr_rubric}

We clarify several details of latent RL training below.
For retrieval feedback, the target candidates follow deterministic mean
trajectories and remain fixed within each rollout group. The retrieval reward
can be written as
\begin{equation}
    R_{\mathrm{ret}}=s_g^{+}-\sum_{j\in\mathcal N}\omega_{g,j}s_{g,j}^{-}.
    \label{eq:reward}
\end{equation}
The negative weights are
$\omega_{g,j}=\operatorname{softmax}_{j\in\mathcal N}(s_{g,j}^{-}/\tau_r)$,
with reward temperature $\tau_r$.
For SDR, the frozen decoder greedily generates one answer per trajectory, and
the fixed evaluator scores reference-answer agreement in $[0,1]$.
For each query, normalize the $G$ rewards to obtain the advantage
$A_g=(R(Z_g)-\bar R)/(\operatorname{std}_{g=1}^{G}[R(Z_g)]+10^{-6})$.
At latent step $k$, the policy ratio is
\begin{equation}
    \rho_{g,k}=\exp\!\left[
       \frac{1}{d}\sum_{j=1}^{d}\log
       \frac{\pi_{\boldsymbol{\theta}}(z_{g,k,j}\mid q,z_{g,<k})}
            {\pi_{\mathrm{old}}(z_{g,k,j}\mid q,z_{g,<k})}\right],
    \label{eq:ratio}
\end{equation}
where $j$ indexes the $d$ action dimensions. To control the numerical scale,
we average the log-density ratio over the action dimensions. The RL objective is
\begin{align}
\mathcal L_{\mathrm{RL}}
={}&\mathbb E_{g,k}\!\Bigl[
    -\min\!\bigl\{\rho_{g,k}A_g,
       \operatorname{clip}(\rho_{g,k},1-\epsilon_-,1+\epsilon_+)A_g\bigr\}
       \notag\\[-1pt]
 &\qquad+\beta\KL\!\left(\pi_{\boldsymbol{\theta}}(\cdot\mid q,z_{g,<k})
                 \Vert\pi_{\mathrm{ref}}(\cdot\mid q,z_{g,<k})\right)
    \Bigr].
\label{eq:grpo}
\end{align}
The clipped term favors higher-reward trajectories, while the KL term keeps the
policy close to the frozen SFT reference $\pi_{\mathrm{ref}}$.
Specifically, our experiments use $G=8$, $\lambda_{\mathrm{SDR}}=0.2$,
$\tau_r=0.2$, $(\epsilon_-,\epsilon_+)=(0.2,0.28)$, and
$\beta=0.01/d$ in Equation~\ref{eq:grpo}.

\noindent\textbf{Semantic decoding reward.}
The SDR component uses a fixed grading contract. The rubric judge $J$ receives the
task question, reference answer $y$, and decoded candidate, and follows a fixed
instruction $\mathcal I_{\mathrm{eval}}$: decompose the reference into
question-relevant atomic facts and score their correctness in the candidate,
accepting paraphrases but penalizing factual contradictions, reversed relations,
and incorrect numerical values, without using external knowledge or rewarding
style. The judge greedily emits a single token constrained to
$\{\texttt{a},\texttt{b},\texttt{c},\texttt{d},\texttt{f}\}$, mapped to the scalar
reward in Table~\ref{tab:sdr_grades}; each latent trajectory is decoded once
without sampling and receives one such score.
\par
\begingroup
\setlength{\intextsep}{\baselineskip}
\begin{table}[h!]
    \centering
    \setlength{\abovecaptionskip}{0pt}
    \caption{SDR grade-to-reward mapping.}
    \label{tab:sdr_grades}
    \vspace{4pt}
    \small
    \begin{tabular}{ccp{0.66\linewidth}}
    \toprule
    Grade & Reward & Criterion \\
    \midrule
    \texttt{a} & $0.00$ & No requested fact is correct, or the sole atomic answer is
    wrong, contradictory, irrelevant, or absent. \\
    \texttt{b} & $0.25$ & Only a peripheral fragment is correct; the requested answer
    is mostly wrong. \\
    \texttt{c} & $0.50$ & A meaningful subset of facts is correct, but another
    substantive fact is wrong or missing. \\
    \texttt{d} & $0.75$ & All core facts are correct, with only a minor non-core defect. \\
    \texttt{f} & $1.00$ & All answer facts are correct, with no contradiction. \\
    \bottomrule
    \end{tabular}
\end{table}
\endgroup

\subsection{Configuration Details}
\label{app:configuration}

As shown in Table~\ref{tab:optimization_recipe}, latent SFT starts from Qwen2-VL
and trains for 3,200 steps with an effective batch size of 512 pairs. Latent RL
then refines the aligned checkpoint for 500 steps on approximately 10K records,
using an effective batch size of 64 queries and a learning rate of $10^{-6}$.
Both stages use $K=8$ latent steps, with RL sampling $G=8$ trajectories per query.

\begin{center}
    \refstepcounter{table}\label{tab:optimization_recipe}
    \begin{minipage}{0.98\textwidth}
    \centering
    \small
    Table~\thetable: Optimization settings for the reported runs.\\[4pt]
    \setlength{\tabcolsep}{6pt}
    \renewcommand{\arraystretch}{1.12}
    \begin{tabular*}{\linewidth}{@{\extracolsep{\fill}}lcc@{}}
        \toprule
        Setting & Latent SFT & Latent RL \\
        \midrule
        Initialization & Qwen2-VL-2B/7B & aligned SFT checkpoint \\
        Training pool & $\sim$1.55M pairs & $\sim$127K candidates \\
        Records used for RL training & -- & $\sim$10K records \\
        Optimizer steps & 3,200 & 500 \\
        Effective batch size & 512 pairs & 64 queries \\
        Latent steps $K$ & 8 & 8 \\
        Trajectories per query $G$ & 1 sampled trajectory & 8 sampled trajectories \\
        Backbone learning rate & $1\times10^{-5}$ & $1\times10^{-6}$ \\
        New-module learning rate & $2\times10^{-5}$ & $1\times10^{-6}$ \\
        Contrastive/reward temperature & $\tau=0.02$ & $\tau_r=0.2$ \\
        Other coefficients & $\lambda_z=\lambda_{\mathrm{ans}}=1$,
            $\lambda_{\mathrm{div}}=0.1$
            & $\lambda_{\mathrm{SDR}}=0.2$ \\
        Policy clipping & -- & $[0.80,1.28]$ \\
        Precision / devices & bfloat16 / 8 GPUs & bfloat16 / 8 GPUs \\
        \bottomrule
    \end{tabular*}
    \end{minipage}
\end{center}

\section{Proof of the Variational Lower Bound}
\label{app:elbo}

This section proves the variational lower bound underlying the latent model,
derives its autoregressive trajectory KL, and relates the result to the
normalized training losses.

\subsection{Conditional Model and Lower Bound}

Fix an input--answer pair $(x,y)$ and the parameters. Given
$Z=(z_1,\ldots,z_K)$, the recurrence deterministically produces
$H_\theta(x,Z)$. Abbreviate the decoder likelihood as
$r(y\mid x,Z)=p_\psi(y\mid\mathcal I_{\mathrm{dec}},H_\theta(x,Z))$.
It is the product of conditional probabilities of the supervised answer tokens,
including retained end tokens; prompt and padding positions are excluded.
For a truncated answer, it denotes the observed-prefix probability.
With $p_0(Z)=\prod_{k=1}^{K}\mathcal N(z_k;0,I)$, define
\begin{equation}
\begin{aligned}
    p_{\theta,\psi}(y,Z\mid x)&=p_0(Z)\,r(y\mid x,Z),\\
    p_{\theta,\psi}(y\mid x)&=\int p_0(Z)\,r(y\mid x,Z)\,dZ.
\end{aligned}
\label{eq:conditional_latent_model}
\end{equation}
The decoder defines a normalized answer distribution and retains access to
$x$ through $H_\theta(x,Z)$. The prior $p_0$ is distinct from the learned
rollout distribution $q(Z)=q_{\boldsymbol{\theta}}(Z\mid x)$ in
Equation~\ref{eq:vae}. Assume finite conditional Gaussian means,
$0<\sigma_{kj}<\infty$, positive evidence, and finite expected negative
log-likelihood and trajectory KL. Then $q$ has a positive density throughout
$\mathbb R^{Kd}$.

\noindent\textbf{Proof.}
Rewriting the marginal as an expectation under $q$ and applying Jensen's
inequality gives
\begin{equation}
\begin{aligned}
    \log p_{\theta,\psi}(y\mid x)
      &=\log\mathbb E_{Z\sim q}\!
          \left[\frac{p_0(Z)r(y\mid x,Z)}{q(Z)}\right]\\
      &\ge\mathbb E_{Z\sim q}
          [\log r(y\mid x,Z)+\log p_0(Z)-\log q(Z)]\\
      &=\mathbb E_{Z\sim q}\log r(y\mid x,Z)
          -\KL(q\Vert p_0)
       \;\equiv\;\mathcal E(x,y).
\end{aligned}
\label{eq:elbo_proof}
\end{equation}
By Bayes' rule,
$p_{\theta,\psi}(Z\mid x,y)=p_0(Z)r(y\mid x,Z)/p_{\theta,\psi}(y\mid x)$.
Substitution into the posterior KL gives the exact gap:
\begin{equation}
    \log p_{\theta,\psi}(y\mid x)-\mathcal E(x,y)
    =\KL\!\left(q(Z)\,\Vert\,p_{\theta,\psi}(Z\mid x,y)\right)\ge0.
\label{eq:elbo_gap}
\end{equation}
Equality holds if and only if $q(Z)=p_{\theta,\psi}(Z\mid x,y)$ almost
everywhere. This proves Equation~\ref{eq:elbo_proof}.\hfill$\square$

The identity holds at each fixed parameter value, including shared $\theta$.
Conditioning $q$ only on $x$ restricts the variational family and may loosen
the bound, but does not invalidate it. No Jacobian of $H_\theta$ is needed:
the integral is over actions $Z$, not a density over hidden states.

\subsection{Autoregressive Trajectory KL}

Let $q_k(\cdot\mid x,z_{<k})=q_{\boldsymbol{\theta}}(\cdot\mid x,z_{<k})$ and
$p_{0,k}=\mathcal N(0,I)$. Expanding the autoregressive density ratio
and applying iterated expectation yields
\begin{equation}
\begin{aligned}
    \KL(q\Vert p_0)
      &=\mathbb E_{Z\sim q}\sum_{k=1}^{K}
          \log\frac{q_k(z_k\mid x,z_{<k})}{p_{0,k}(z_k)}\\
      &=\sum_{k=1}^{K}\mathbb E_{z_{<k}\sim q}\!
          \left[\mathbb E_{z_k\sim q_k(\cdot\mid x,z_{<k})}
          \log\frac{q_k(z_k\mid x,z_{<k})}{p_{0,k}(z_k)}\right]\\
      &=\sum_{k=1}^{K}\mathbb E_{z_{<k}\sim q}
          \KL\!\left(q_k(\cdot\mid x,z_{<k})\Vert p_{0,k}\right).
\end{aligned}
\label{eq:trajectory_kl_chain}
\end{equation}
Future actions integrate out to one; the prefix at $k=1$ is empty.
These are expected conditional KLs, not KLs of the marginal action distributions.
At a fixed prefix, the Gaussian moments
$\mathbb E[(z_{kj}-\mu_{kj})^2]=\sigma_{kj}^2$ and
$\mathbb E[z_{kj}^2]=\mu_{kj}^2+\sigma_{kj}^2$ give
\begin{equation}
    \KL(q_k\Vert p_{0,k})
    =\frac12\sum_{j=1}^{d}
       \left(\mu_{kj}^2+\sigma_{kj}^2-1-2\log\sigma_{kj}\right).
\label{eq:elbo_gaussian_kl}
\end{equation}
Consequently, Equations~\ref{eq:kl_main} and~\ref{eq:kl} imply
\begin{equation}
    \KL(q\Vert p_0)
    =Kd\,\mathbb E_{Z\sim q}\mathcal L_{\mathrm{KL}}(x,Z).
\label{eq:trajectory_kl_estimator}
\end{equation}
Although each conditional KL is analytic, its parameters depend on the sampled
prefix. A single-rollout $\mathcal L_{\mathrm{KL}}$ is therefore an unbiased
estimate of the trajectory KL divided by $Kd$. Likewise,
$\log r(y\mid x,Z)-Kd\,\mathcal L_{\mathrm{KL}}(x,Z)$ has expectation
$\mathcal E(x,y)$; an individual sampled value need not be a lower bound.

\subsection{Relation to the Normalized Training Losses}

Let $\mathcal L_{\mathrm{ans}}^{\mathrm{seq}}$ denote the expected
negative log-likelihood per supervised answer sequence in
Equation~\ref{eq:decode_main}. Averaging the bound gives
\begin{equation}
    -\mathbb E_{(x,y)}\mathcal E(x,y)
    =\mathcal L_{\mathrm{ans}}^{\mathrm{seq}}
     +Kd\,\mathbb E_{x,\,Z\sim q_{\boldsymbol{\theta}}(\cdot\mid x)}
       \mathcal L_{\mathrm{KL}}(x,Z).
\label{eq:negative_elbo_losses}
\end{equation}
For a fixed supervised-side batch $\mathcal B_s$, write
$m_s=|\mathcal B_s|>0$, $N_s>0$ for its valid-token count, and
$\ell_s=N_s/m_s$. Define the per-example averages
$\overline{\mathcal E}_s=m_s^{-1}\sum_{i\in\mathcal B_s}\mathcal E(x_i,y_i)$
and $\mathcal L_{\mathrm{KL}}^{(s)}
=m_s^{-1}\sum_{i\in\mathcal B_s}\mathcal L_{\mathrm{KL}}(x_i,Z_i)$.
The token loss in Equation~\ref{eq:decode} divides the total answer negative
log-likelihood by $N_s$ rather than $m_s$, so
\begin{equation}
    -\overline{\mathcal E}_s
    =\ell_s\,\mathbb E_Z\mathcal L_{\mathrm{ans}}^{(s)}
     +Kd\,\mathbb E_Z\mathcal L_{\mathrm{KL}}^{(s)}.
\label{eq:elbo_side_normalization}
\end{equation}
This identity is exact even for unequal answer lengths. For multiple
supervised sides, average the right-hand side over sides, retaining each
$\ell_s$. When lengths vary across batches, this factor must remain inside
the batch expectation. Answer and KL terms must cover the same examples and
sides; KL on an unsupervised side is additional regularization.

For $\lambda_{\mathrm{ans}}>0$, matching the answer-plus-KL objective to
$-\overline{\mathcal E}_s$ up to a positive scale requires
\begin{equation}
    \frac{\lambda_{\mathrm{KL}}}{\lambda_{\mathrm{ans}}}
      =\frac{Kd}{\ell_s},\qquad
    \mathbb E_Z\!\left[
      \lambda_{\mathrm{ans}}\mathcal L_{\mathrm{ans}}^{(s)}
      +\lambda_{\mathrm{KL}}\mathcal L_{\mathrm{KL}}^{(s)}\right]
      =-\frac{\lambda_{\mathrm{ans}}}{\ell_s}\overline{\mathcal E}_s.
\label{eq:elbo_weight_matching}
\end{equation}
A fixed ratio cannot generally match this scaling across sides or batches
with different $\ell_s$. Equation~\ref{eq:sft} independently weights the
normalized losses and adds retrieval and diversity supervision. Thus the
proof supplies a variational interpretation of the answer and KL terms,
without equating the complete SFT objective to the standard negative ELBO.
The bound concerns stochastic trajectories; deterministic mean-path inference
is not the marginalization in Equation~\ref{eq:conditional_latent_model}.

%% file: references.bib
@article{rime,
  title={{Beyond Chain-of-Thought: Rewrite as a Universal Interface for Generative Multimodal Embeddings}},
  author={Wu, Peixi and Mei, Ke and Ma, Feipeng and Chai, Bosong and Lan, Zhibin and Zhao, Chenxi and Yan, Shannan and Chen, Jie and Hu, Zhangchi and Peng, Yansong and Lin, Bo and Zhou, Junjie and Yin, Dacheng and Wang, Tianyi and Rao, Fengyun and Lyu, Jing and Li, Hebei and Sun, Xiaoyan},
  journal={arXiv preprint arXiv:2604.22280},
  year={2026},
  url={https://arxiv.org/abs/2604.22280}
}

@article{lame,
  title={{LaME: Learning to Think in Latent Space for Multimodal Embedding via Information Bottleneck}},
  author={Wu, Peixi and Yang, Biao and Ma, Feipeng and Chai, Bosong and Lin, Bo and Yuan, Wei and Yang, Fan and Gao, Tingting and Li, Hebei and Sun, Xiaoyan},
  journal={arXiv preprint arXiv:2606.13061},
  year={2026},
  url={https://arxiv.org/abs/2606.13061}
}

@inproceedings{vlm2vec,
  title={{VLM2Vec: Training Vision-Language Models for Massive Multimodal Embedding Tasks}},
  author={Jiang, Ziyan and Meng, Rui and Yang, Xinyi and Yavuz, Semih and Zhou, Yingbo and Chen, Wenhu},
  booktitle={The Thirteenth International Conference on Learning Representations},
  year={2025},
  url={https://openreview.net/forum?id=TE0KOzWYAF}
}

@article{vlm2vecv2,
  title={{VLM2Vec-V2: Advancing Multimodal Embedding for Videos, Images, and Visual Documents}},
  author={Meng, Rui and Jiang, Ziyan and Liu, Ye and Su, Mingyi and Yang, Xinyi and Fu, Yuepeng and Qin, Can and Thirukovalluru, Raghuveer and Zhang, Xuan and Chen, Zeyuan and Xu, Ran and Xiong, Caiming and Zhou, Yingbo and Chen, Wenhu and Yavuz, Semih},
  journal={Transactions on Machine Learning Research},
  year={2026},
  url={https://openreview.net/forum?id=TpU38jbKIJ}
}

@inproceedings{gme,
  title={{Bridging Modalities: Improving Universal Multimodal Retrieval by Multimodal Large Language Models}},
  author={Zhang, Xin and Zhang, Yanzhao and Xie, Wen and Li, Mingxin and Dai, Ziqi and Long, Dingkun and Xie, Pengjun and Zhang, Meishan and Li, Wenjie and Zhang, Min},
  booktitle={Proceedings of the IEEE/CVF Conference on Computer Vision and Pattern Recognition},
  pages={9274--9285},
  year={2025},
  doi={10.1109/CVPR52734.2025.00866},
  url={https://doi.org/10.1109/CVPR52734.2025.00866}
}

@inproceedings{mmembed,
  title={{MM-Embed: Universal Multimodal Retrieval with Multimodal LLMs}},
  author={Lin, Sheng-Chieh and Lee, Chankyu and Shoeybi, Mohammad and Lin, Jimmy and Catanzaro, Bryan and Ping, Wei},
  booktitle={The Thirteenth International Conference on Learning Representations},
  year={2025},
  url={https://openreview.net/forum?id=i45NQb2iKO}
}

@article{metaembed,
  title={{MetaEmbed: Scaling Multimodal Retrieval at Test-Time with Flexible Late Interaction}},
  author={Xiao, Zilin and Ma, Qi and Gu, Mengting and Chen, Chun-cheng Jason and Chen, Xintao and Ordonez, Vicente and Mohan, Vijai},
  journal={arXiv preprint arXiv:2509.18095},
  year={2026},
  url={https://arxiv.org/abs/2509.18095}
}

@inproceedings{coconut,
  title={Training Large Language Models to Reason in a Continuous Latent Space},
  author={Hao, Shibo and Sukhbaatar, Sainbayar and Su, DiJia and Li, Xian and Hu, Zhiting and Weston, Jason and Tian, Yuandong},
  booktitle={The Second Conference on Language Modeling},
  year={2025},
  url={https://openreview.net/forum?id=Itxz7S4Ip3}
}

@article{embedrl,
  title={{Embed-RL}: Reinforcement Learning for Reasoning-Driven Multimodal Embeddings},
  author={Jiang, Haonan and Wang, Yuji and Zhu, Yongjie and Lu, Xin and Qin, Wenyu and Wang, Meng and Wan, Pengfei and Tang, Yansong},
  journal={arXiv preprint arXiv:2602.13823},
  year={2026},
  url={https://arxiv.org/abs/2602.13823}
}

@inproceedings{vib,
  title={Deep Variational Information Bottleneck},
  author={Alemi, Alexander A. and Fischer, Ian and Dillon, Joshua V. and Murphy, Kevin},
  booktitle={Proceedings of the 5th International Conference on Learning Representations (ICLR)},
  year={2017}
}

@article{mrmr,
  title={{MRMR}: A Realistic and Expert-Level Multidisciplinary Benchmark for Reasoning-Intensive Multimodal Retrieval},
  author={Zhang, Siyue and Gao, Yuan and Zhou, Xiao and Zhao, Yilun and Song, Tingyu and Cohan, Arman and Luu, Anh Tuan and Zhao, Chen},
  journal={arXiv preprint arXiv:2510.09510},
  year={2025},
  url={https://arxiv.org/abs/2510.09510}
}

@article{qwen2vl,
  title={Qwen2-vl: Enhancing vision-language model's perception of the world at any resolution},
  author={Wang, Peng and Bai, Shuai and Tan, Sinan and Wang, Shijie and Fan, Zhihao and Bai, Jinze and Chen, Keqin and Liu, Xuejing and Wang, Jialin and Ge, Wenbin and others},
  journal={arXiv preprint arXiv:2409.12191},
  year={2024}
}

@article{tteflash,
  title={{TTE-Flash}: Accelerating Reasoning-based Multimodal Representations via Think-Then-Embed Tokens},
  author={Cheng, Jianpeng and Wu, Xian and Zhang, Jiangfan and Bao, Wentao and Ahuja, Chaitanya and Mishra, Shlok Kumar and Yu, Hanchao and Gao, Yang and Xia, Fan and Guo, Qi and Zhai, Shaodan and Fan, Xiangjun and Xiao, Jun},
  journal={arXiv preprint arXiv:2605.16638},
  year={2026},
  url={https://arxiv.org/abs/2605.16638v1}
}

@article{vae,
  title={Auto-Encoding Variational Bayes},
  author={Kingma, Diederik P. and Welling, Max},
  journal={arXiv preprint arXiv:1312.6114},
  year={2013},
  url={https://arxiv.org/abs/1312.6114}
}

@article{grpo,
  title={{DeepSeekMath}: Pushing the Limits of Mathematical Reasoning in Open Language Models},
  author={Shao, Zhihong and Wang, Peiyi and Zhu, Qihao and Xu, Runxin and Song, Junxiao and Bi, Xiao and Zhang, Haowei and Zhang, Mingchuan and Li, Y. K. and Wu, Y. and Guo, Daya},
  journal={arXiv preprint arXiv:2402.03300},
  year={2024},
  url={https://arxiv.org/abs/2402.03300}
}

@inproceedings{clip,
  title={Learning Transferable Visual Models From Natural Language Supervision},
  author={Radford, Alec and Kim, Jong Wook and Hallacy, Chris and Ramesh, Aditya and Goh, Gabriel and Agarwal, Sandhini and Sastry, Girish and Askell, Amanda and Mishkin, Pamela and Clark, Jack and Krueger, Gretchen and Sutskever, Ilya},
  booktitle={Proceedings of the 38th International Conference on Machine Learning},
  series={Proceedings of Machine Learning Research},
  volume={139},
  pages={8748--8763},
  year={2021},
  publisher={PMLR},
  url={https://proceedings.mlr.press/v139/radford21a.html}
}

@article{uniir,
  title={{UniIR}: Training and Benchmarking Universal Multimodal Information Retrievers},
  author={Wei, Cong and Chen, Yang and Chen, Haonan and Hu, Hexiang and Zhang, Ge and Fu, Jie and Ritter, Alan and Chen, Wenhu},
  journal={arXiv preprint arXiv:2311.17136},
  year={2023},
  url={https://arxiv.org/abs/2311.17136}
}

@article{e5v,
  title={{E5-V}: Universal Embeddings with Multimodal Large Language Models},
  author={Jiang, Ting and Song, Minghui and Zhang, Zihan and Huang, Haizhen and Deng, Weiwei and Sun, Feng and Zhang, Qi and Wang, Deqing and Zhuang, Fuzhen},
  journal={arXiv preprint arXiv:2407.12580},
  year={2024},
  url={https://arxiv.org/abs/2407.12580}
}

@article{llave,
  title={{LLaVE}: Large Language and Vision Embedding Models with Hardness-Weighted Contrastive Learning},
  author={Lan, Zhibin and Niu, Liqiang and Meng, Fandong and Zhou, Jie and Su, Jinsong},
  journal={arXiv preprint arXiv:2503.04812},
  year={2025},
  url={https://arxiv.org/abs/2503.04812}
}

@article{umer1,
  title={{UME-R1}: Exploring Reasoning-Driven Generative Multimodal Embeddings},
  author={Lan, Zhibin and Niu, Liqiang and Meng, Fandong and Zhou, Jie and Su, Jinsong},
  journal={arXiv preprint arXiv:2511.00405},
  year={2025},
  url={https://arxiv.org/abs/2511.00405}
}

@article{tte,
  title={Think Then Embed: Generative Context Improves Multimodal Embedding},
  author={Cui, Xuanming and Cheng, Jianpeng and Chen, Hong-you and Shukla, Satya Narayan and Awasthi, Abhijeet and Pan, Xichen and Ahuja, Chaitanya and Mishra, Shlok Kumar and Yang, Yonghuan and Xiao, Jun and Guo, Qi and Lim, Ser-Nam and Singh, Aashu and Fan, Xiangjun},
  journal={arXiv preprint arXiv:2510.05014},
  year={2025},
  url={https://arxiv.org/abs/2510.05014}
}

@article{plume,
  title={{PLUME}: Latent Reasoning Based Universal Multimodal Embedding},
  author={He, Chenwei and Hao, Xiangzhao and Yang, Tianyu and Ma, Yuxiang and Jia, Yuheng and Wu, Lingxiang and Zhao, Chaoyang and Guo, Haiyun and Wang, Jinqiao},
  journal={arXiv preprint arXiv:2604.02073},
  year={2026},
  url={https://arxiv.org/abs/2604.02073}
}

@article{mmembr1,
  title={{MMEmb-R1}: Reasoning-Enhanced Multimodal Embedding with Pair-Aware Selection and Adaptive Control},
  author={Wang, Yuchi and Yang, Dingkang and Yu, Haiyang and Bian, Weikang and Long, Jiefeng and Liang, Xiao and Feng, Chao and Li, Hongsheng},
  journal={arXiv preprint arXiv:2604.06156},
  year={2026},
  url={https://arxiv.org/abs/2604.06156}
}

@article{codi,
  title={{CODI}: Compressing Chain-of-Thought into Continuous Space via Self-Distillation},
  author={Shen, Zhenyi and Yan, Hanqi and Zhang, Linhai and Hu, Zhanghao and Du, Yali and He, Yulan},
  journal={arXiv preprint arXiv:2502.21074},
  year={2025},
  url={https://arxiv.org/abs/2502.21074}
}

@article{softcotpp,
  title={{SoftCoT++}: Test-Time Scaling with Soft Chain-of-Thought Reasoning},
  author={Xu, Yige and Guo, Xu and Zeng, Zhiwei and Miao, Chunyan},
  journal={arXiv preprint arXiv:2505.11484},
  year={2025},
  url={https://arxiv.org/abs/2505.11484}
}

@inproceedings{rezende,
  title={Stochastic Backpropagation and Approximate Inference in Deep Generative Models},
  author={Rezende, Danilo Jimenez and Mohamed, Shakir and Wierstra, Daan},
  booktitle={Proceedings of the 31st International Conference on Machine Learning},
  series={Proceedings of Machine Learning Research},
  volume={32},
  pages={1278--1286},
  year={2014},
  publisher={PMLR},
  url={https://proceedings.mlr.press/v32/rezende14.html}
}

@article{laser,
  title={{LaSER}: Internalizing Explicit Reasoning into Latent Space for Dense Retrieval},
  author={Jin, Jiajie and Zhang, Yanzhao and Li, Mingxin and Long, Dingkun and Xie, Pengjun and Zhu, Yutao and Dou, Zhicheng},
  journal={arXiv preprint arXiv:2603.01425},
  year={2026},
  url={https://arxiv.org/abs/2603.01425}
}

@inproceedings{unime,
  title={Breaking the Modality Barrier: Universal Embedding Learning with Multimodal {LLMs}},
  author={Gu, Tiancheng and Yang, Kaicheng and Feng, Ziyong and Wang, Xingjun and Zhang, Yanzhao and Long, Dingkun and Chen, Yingda and Cai, Weidong and Deng, Jiankang},
  booktitle={Proceedings of the 33rd ACM International Conference on Multimedia},
  year={2025}
}

@article{rzenembed,
  title={{RzenEmbed}: Towards Comprehensive Multimodal Retrieval},
  author={Jian, Weijian and Zhang, Yajun and Liang, Dawei and Xie, Chunyu and He, Yixiao and Leng, Dawei and Yin, Yuhui},
  journal={arXiv preprint arXiv:2510.27350},
  year={2025},
  url={https://arxiv.org/abs/2510.27350}
}

@inproceedings{colbertv2,
  title={{ColBERTv2}: Effective and Efficient Retrieval via Lightweight Late Interaction},
  author={Santhanam, Keshav and Khattab, Omar and Saad-Falcon, Jon and Potts, Christopher and Zaharia, Matei},
  booktitle={Proceedings of the 2022 Conference of the North American Chapter of the Association for Computational Linguistics: Human Language Technologies},
  pages={3715--3734},
  year={2022}
}

@article{implicitcot,
  title={Implicit Chain of Thought Reasoning via Knowledge Distillation},
  author={Deng, Yuntian and Prasad, Kiran and Fernandez, Roland and Smolensky, Paul and Chaudhary, Vishrav and Shieber, Stuart},
  journal={arXiv preprint arXiv:2311.01460},
  year={2023},
  url={https://arxiv.org/abs/2311.01460}
}

@inproceedings{softcot,
  title={{SoftCoT}: Soft Chain-of-Thought for Efficient Reasoning with {LLMs}},
  author={Xu, Yige and Guo, Xu and Zeng, Zhiwei and Miao, Chunyan},
  booktitle={Proceedings of the 63rd Annual Meeting of the Association for Computational Linguistics},
  year={2025}
}

@inproceedings{omib,
  title={Learning Optimal Multimodal Information Bottleneck Representations},
  author={Wu, Qilong and Shao, Yiyang and Wang, Jun and Sun, Xiaobo},
  booktitle={Proceedings of the 42nd International Conference on Machine Learning},
  year={2025}
}

@article{ib,
  title={The Information Bottleneck Method},
  author={Tishby, Naftali and Pereira, Fernando C. and Bialek, William},
  journal={arXiv preprint physics/0004057},
  year={2000},
  url={https://arxiv.org/abs/physics/0004057}
}

@inproceedings{lamra,
  title={{LamRA}: Large Multimodal Model as Your Advanced Retrieval Assistant},
  author={Liu, Yikun and Zhang, Yajie and Cai, Jiayin and Jiang, Xiaolong and Hu, Yao and Yao, Jiangchao and Wang, Yanfeng and Xie, Weidi},
  booktitle={Proceedings of the IEEE/CVF Conference on Computer Vision and Pattern Recognition},
  pages={4015--4025},
  year={2025},
  url={https://openaccess.thecvf.com/content/CVPR2025/html/Liu_LamRA_Large_Multimodal_Model_as_Your_Advanced_Retrieval_Assistant_CVPR_2025_paper.html}
}

@inproceedings{mme5,
  title={{mmE5}: Improving Multimodal Multilingual Embeddings via High-quality Synthetic Data},
  author={Chen, Haonan and Wang, Liang and Yang, Nan and Zhu, Yutao and Zhao, Ziliang and Wei, Furu and Dou, Zhicheng},
  booktitle={Findings of the Association for Computational Linguistics: ACL 2025},
  pages={8254--8275},
  year={2025},
  doi={10.18653/v1/2025.findings-acl.433},
  url={https://aclanthology.org/2025.findings-acl.433/}
}

@inproceedings{cafe,
  title={{CAFe}: Unifying Representation and Generation with Contrastive-Autoregressive Finetuning},
  author={Yu, Hao and Zhao, Zhuokai and Yan, Shen and Korycki, Lukasz and Wang, Jianyu and He, Baosheng and Liu, Jiayi and Zhang, Lizhu and Fan, Xiangjun and Yu, Hanchao},
  booktitle={Proceedings of the IEEE/CVF International Conference on Computer Vision Workshops},
  pages={6345--6356},
  year={2025},
  url={https://openaccess.thecvf.com/content/ICCV2025W/Findings/html/Yu_CAFE_Unifying_Representation_and_Generation_with_Contrastive-Autoregressive_Finetuning_ICCVW_2025_paper.html}
}

@inproceedings{colar,
  title={Think Silently, Think Fast: Dynamic Latent Compression of {LLM} Reasoning Chains},
  author={Tan, Wenhui and Li, Jiaze and Ju, Jianzhong and Luo, Zhenbo and Song, Ruihua and Luan, Jian},
  booktitle={Advances in Neural Information Processing Systems},
  volume={38},
  year={2025},
  url={https://proceedings.neurips.cc/paper_files/paper/2025/hash/0706261aedab63814a2b73c32564b4c4-Abstract-Conference.html}
}

@article{climb,
  title={Contrastive Learning via Variational Information Bottleneck},
  author={Li, Jin and Wang, Yaoming and Zhang, Xiaopeng and Jiang, Dongsheng and Dai, Wenrui and Li, Chenglin and Xiong, Hongkai and Tian, Qi},
  journal={IEEE Transactions on Pattern Analysis and Machine Intelligence},
  volume={47},
  number={9},
  pages={7410--7427},
  year={2025},
  doi={10.1109/TPAMI.2025.3571990}
}

@article{tsembed,
  title={{TSEmbed}: Unlocking Task Scaling in Universal Multimodal Embeddings},
  author={Wu, Yebo and Liu, Feng and Xie, Ziwei and Liu, Zhiyuan and Zhang, Changwang and Wang, Jun and Li, Li},
  journal={arXiv preprint arXiv:2603.04772},
  year={2026},
  url={https://arxiv.org/abs/2603.04772}
}

@inproceedings{unimev2,
  title={{UniME-V2}: {MLLM}-as-a-Judge for Universal Multimodal Embedding Learning},
  author={Gu, Tiancheng and Yang, Kaicheng and Zhang, Kaichen and An, Xiang and Feng, Ziyong and Zhang, Yueyi and Cai, Weidong and Deng, Jiankang and Bing, Lidong},
  booktitle={Proceedings of the AAAI Conference on Artificial Intelligence},
  volume={40},
  pages={21378--21386},
  year={2026},
  doi={10.1609/aaai.v40i26.39284},
  url={https://ojs.aaai.org/index.php/AAAI/article/view/39284}
}
